\documentclass[12pt,a4paper]{article}
\usepackage[margin=1in]{geometry}
\usepackage[utf8]{inputenc}
\usepackage[T1]{fontenc}
\usepackage{tgtermes}
\usepackage{graphicx, float}
\usepackage{booktabs}
\usepackage{siunitx}
\usepackage{dcolumn, multirow}
\usepackage{amsfonts, amsmath, amssymb, bm, mathtools}
\usepackage{xcolor}
\usepackage{url}
\usepackage{threeparttable}
\usepackage{csquotes}
\usepackage{rotating}
\usepackage{tikz}
\usepackage[labelsep=colon]{caption}
\usepackage[font=footnotesize]{subcaption}
\usepackage{natbib}
\usepackage{parskip}
\usepackage{enumitem}
\usepackage{longtable}
\usepackage{tabularx}
\usepackage{array}
\newcolumntype{Y}{>{\raggedright\arraybackslash}X}
\usepackage{setspace}
\usepackage[bottom]{footmisc}
\usepackage{titlesec}
    \titleformat{\section}{\normalsize\bfseries}{\thesection}{1em}{}
    \titleformat{\subsection}{\small\it}{}{1em}{}
    \titleformat{\subsubsection}[runin]{\small\it}{}{0em}{}[.\quad]
\usepackage[hidelinks]{hyperref}

\renewcommand{\thesection}{5.\arabic{section}}

\newcommand{\bookrepo}{\url{https://github.com/llms-political-social-science-book-oup/book-code-data}}

\title{\large\bfseries Replicability, Reproducibility, and Robustness in\\ Political and Social Science Research with Large Language Models\thanks{Cross-references to other chapters, sections, and tables refer to the full book. Data and code: \bookrepo.}}
\author{Joan C. Timoneda\thanks{Assistant Professor, Department of Political Science, Purdue University. Email: timoneda@purdue.edu.} \and Sebasti\'an Vallejo Vera\thanks{Assistant Professor, Department of Political Science, University of Western Ontario.}}
\date{\normalsize Chapter 5 in \textit{LLMs in Political and Social Science: Principles\\ and Applications of Transformer Models}.\\[6pt] Forthcoming in 2027, Oxford University Press.}

\begin{document}
\setstretch{1.2}
\maketitle

\begin{abstract}
\noindent This chapter addresses the central methodological problem of research with large language models: their output is probabilistic, varies across models and versions, and can change downstream statistical results. It distinguishes replicability, reproducibility, and robustness, and proposes a probabilistic standard under which a replication succeeds when it recovers the same distribution of results rather than identical output. It introduces a standardized prompt script, tests within-model variation for encoders and decoders, and develops error-structure analysis to diagnose where models err and whether their errors align. Using three datasets, the chapter shows that measurement choice can flip the sign of downstream regression coefficients. Existing design-based corrections with gold-labeled subsamples restore validity, but come at the cost of estimate precision. The chapter proposes a stacked ensemble of open-weight models that recovers much of the precision lost in the correction step. The chapter concludes with disclosure standards covering code, prompts, codebooks, training data, and trained models.

\end{abstract}

\clearpage


In the earlier chapters of the book, we have laid out clear pipelines for applying encoder and decoder LLMs to annotate, classify, summarize, and recover topics from political texts at scale. We have done so with costs and efficiency in mind, consistently highlighting the benefits of encoder models and open-weight decoders in many applied tasks. Transformer models, especially decoders, come with a series of limitations, several of which we have already touched on in earlier chapters. A subset of such limitations falls within the discipline's growing debate around the \textit{reproducibility} and \textit{transparency} of results in applied LLM research (hereafter RT) \citep{spirling2023open, barrie2024replication, alvarez2022not}.

Let us begin by defining the exact terminology within the RT debate, as the terms are often used interchangeably and inconsistently in the applied literature. Above, we used \textit{reproducibility} and \textit{transparency} as two umbrella terms to refer to the general debate. There are, however, two more important concepts to introduce. Following \citet{barrie2024replication}, we use \textit{replicability} to mean that the same data, code, and model yield the same reported results. A related term we bring to the debate more explicitly is \textit{robustness}, where the same data analyzed under different reasonable LLM choices, such as different models, versions, or prompts, yields similar findings.\footnote{Some disciplines and authorities swap the definitions of reproducibility and replicability \citep[see][]{national2019reproducibility}.} Although \citet{barrie2024replication} discuss robustness within their broader account of replication, we treat it as a separate dimension. Replicability asks whether we can generate the same results from the same data and model, whereas robustness asks whether we can reach similar results with the same data but different LLMs. Lastly, \textit{reproducibility} in technical terms means that the same model applied to new data produces similar findings. \textit{Transparency} means that an outside observer can understand and verify all the moving parts of an analysis (models, data, and methodology). Therefore, the difference between the terms lies in the component that we hold constant: the model (reproducibility), the data (robustness), or both (replicability). In this chapter, we cover all of these concepts under the overall umbrella of the RT debate. 

At the heart of the issue is that the same complex architecture that makes decoder LLMs so powerful also makes them unpredictable in ways researchers often cannot control or, with closed models, even inspect. It is difficult to guarantee that an LLM result today will be replicated or reproduced tomorrow. And while the problem is real, neither the diagnosis nor the solutions are settled. RT failures are not uniform across all uses of LLMs, and critics of LLMs often hold these models to a higher standard that traditional, well-established methods rarely meet. The constructive task, as we see it, is to identify the distinct problems that threaten reproducible research, be clear about which of them we can fix or avoid, and which we can only control for and report. Our goal in this chapter is to evaluate the current RT debate, engage with it directly, and provide concrete solutions to each of the central issues. 

While RT presents uniquely difficult challenges in the context of Transformer LLM research, we think there are equally strong solutions. Our first contribution is to seek to push the field toward \textit{probabilistic} RT. Until now, many computational pipelines in political science and the social sciences have been relatively \textit{deterministic}: replicating work should return identical results and reproducing it should yield roughly equal ones.\footnote{Even when there are stochastic elements, such as bootstrapping, random forest, MCMC, and BoW topic models, researchers have been able to fix the random elements, often using seeds.} This determinism held even though inference itself has always been probabilistic. Most of this conventional thinking around RT with quantitative approaches collapses with LLMs, as they cannot guarantee deterministic results. Therefore, since Transformer LLMs are inherently probabilistic, we contend that it is in our best interest as social scientists to switch to probabilistic RT. Instead of drawing exact results or expecting close approximations, we should focus on \textit{distributions} of results, leveraging the uncertainty of LLMs to determine whether results are replicable, reproducible and robust. 

In our probabilistic RT framework, we separate \textit{measurement} from \textit{downstream analysis}. There are two reasons for making this conceptual distinction. First, text is an unstructured source of data, and LLMs are often used to extract structured data from it. This is true for annotation tasks but also topic modeling and other tasks. How we approach measurement with LLMs matters \citep{timoneda2025bert}. Second, as researchers, we ultimately care the most about downstream analyses. LLM measures should be good proxies for the concept we want to capture and thus produce unbiased coefficients in our statistical models. This is a focus of recent work, and rightly so \citep{egami2024}. These two aspects of an LLM-focused pipeline, measurement and downstream analysis, are both crucial yet they demand different RT strategies. We provide further detail and solutions along both dimensions in the following sections.

We would also like to frame the RT discussion on LLMs in more general and perhaps more optimistic terms. The first step is to place the problem of RT in a broader discipline-wide context. The reproducibility problem affects many methods, not just LLMs. LLMs do introduce additional sources of variation that complicate reproducibility further. Yet the problem is structural. As a discipline, we are concerned with the reliability of a measure and how close it is to the ground truth (however we define this), but determining the ground truth has generally been difficult and inconsistent. Long before LLMs, when social scientists could only turn text into data by hand, two trained coders would produce different labels from the same codebook. We could take this a step further and say that the same coder, returning a year later, would likely label some texts differently. Coders, in the end, diverge on how they apply rules and determine ambiguous cases. They also bring different background knowledge to the task. 

The same problem surfaces in machine learning, where the benchmark datasets used to evaluate models are themselves human-labeled and contain errors. In fact, \citet{northcutt2021pervasive} audited ten of the most widely used benchmark test sets and found pervasive labeling errors, averaging at least 3.3 percent, with higher rates in some datasets.\footnote{The audited datasets include ImageNet and the IMDB sentiment corpus, among others.} These errors are enough to change the evaluation of which models perform best. Many benchmark datasets are crowdsourced, a practice that builds on research showing that aggregate crowd labels can match expert assessments \citep[for political science, see][]{benoit2016crowd}. If human annotation were reproducible, similar well-designed crowdsourcing procedures should yield similar labels in the aggregate. But this is not always the case. Coders apply different heuristics when presented with multiple examples of correct labels, producing inconsistent labels \citep[see][]{ennser2018impact}. The quality of the coders affects the quality of the labels, in predictable ways as long as quality can be measured, as does their domain expertise. 

Therefore, we suggest thinking about reproducibility of LLM-derived research in light of the general complexities around labeled data, human or otherwise. As \citet{barrie2024replication} and others argue, and we agree, LLMs introduce new challenges that exacerbate the reproducibility problem. Yet we consider these challenges surmountable, requiring a deep understanding of the LLM landscape, the capabilities and flaws of the models, and a clear set of solutions. Our solution rests on a thread that runs throughout this book: for each task, we need a \textit{complete picture} of LLM performance and how it compares with expert coders. Because model performance and error structures are highly task-specific, comparing the output of one or two models will not suffice. Researchers need to apply a theoretically justifiable and varied \textit{ensemble} of LLMs to more comprehensively understand LLM task-specific performance. We provide a full set of guidelines for this in the book, which we call benchmark-compare-harmonize, or BCH, introduced in Section 4.11. BCH covers both halves of the RT problem, measurement quality and the consistency of downstream estimates.

In this chapter, we first discuss the problem of RT in more detail, following \citet{barrie2024replication}. Next, we introduce probabilistic RT and provide a fixed set of prompting strategies and examples for different applied tasks. We then discuss replicability, reproducibility and robustness in different sections and propose solutions to the central problems. We close by harmonizing LLM measures with downstream analyses and discussing transparency in research.

\section{RT: The Achilles' Heel of Applied LLM Research} 

\citet{barrie2024replication} provide one of the most systematic problematizations of LLM research in the social sciences. One of their key concerns is replicability, and they identify two sources of error that affect it: \textit{within}-model and \textit{across}-model variation. They organize coding practices along two questions: whether exact replicability is possible, and whether the approach is \textit{robust} and thus independent of the particular system used to produce it.\footnote{\citet{barrie2024replication} define their second question as \textit{fragility} of replication, or when replication is dependent on access to a particular model or system. We have changed the terminology and use \textit{robustness}, but it captures that same idea.} The classical ``data and code'' replication from \citet{king1995replication} demands that results be regenerated exactly but it is fragile, as it requires the replicator to have the same computational environment as the original researcher. \citet{benoit2016crowd} develop a crowdsourced coding approach that, by contrast, is not exactly replicable (one cannot recruit the identical workers each time). Yet it is robust, as the aggregate annotations across different coders and platforms approximate expert judgment. On \citeauthor{barrie2024replication}'s (\citeyear{barrie2024replication}) account, proprietary LLMs occupy an uncomfortable position: they are neither replicable nor robust. They are not replicable because their output cannot be regenerated exactly. They are not robust because a model version can be updated or simply cease to exist, so even the tool itself may become unavailable. And LLM results also vary significantly across different models and versions \citep{walker2025, timoneda2026auditing}. Results, therefore, are too dependent on the particular tool used in the research.

\citet{barrie2024replication} document both error sources empirically. For within-model variation, they re-run the same labeling tasks with the same model over several months, and separately repeat a task many times and average the output. Both exhibit substantial variation, with important effects on downstream estimates and statistical significance. For across-model variation, they run the same task across different models and versions, again finding sizable differences in output. Making a strong case for transparent and auditable tools, \citet{spirling2023open} advocates for open-weight models, as they provide the smallest within-model variation and allow for model versions to be saved and reused over time \citep{barrie2024replication}. We agree with most of this careful and thoughtful diagnosis.

We would only qualify one part of it: how the authors conceptualize and measure \textit{robustness}. \citet{barrie2024replication} contrast robust crowdsourcing with fragile LLMs by running both over repeated rounds across several months. Run-to-run variability is wider for the language models than for the human coders, except in the case of their locally installed, open-weight Llama implementation, which shows lower variation than the humans. However, this comparison leaves two issues open. First, the evidence points to open-weight models being more robust, and this is a finding the authors themselves highlight. But it also cuts against their own framing. If open-weight LLMs are more robust, then this might be less a general LLM issue, and more a proprietary versus open-weight one. The authors do recommend the use of open-weight models as the closest available option to replication. We take this a step further and make open-weight models central to the study, use, and improvement of LLMs in social science research.

Our second qualification is more conceptual. \citeauthor{barrie2024replication}'s (\citeyear{barrie2024replication}) comparison is asymmetric. On the LLM side, they measure variation across several different models, but on the human side they measure it within a single pool of crowd workers --- the pool they characterize as robust. This is an informative comparison, but a single pool will naturally show less variation than a set of distinct models. The apples-to-apples comparison would set multiple models against multiple pools of coders. We do not actually know whether the human side would look as stable if the comparison were built symmetrically. Would the variance be the same, lower, or higher if we re-ran the task each month with three fresh undergraduate or graduate research assistants? Or with assistants we trained ourselves rather than others working only from written instructions? Robustness requires us to ask \textit{which} model and \textit{which} human, and to match the sources of variation on each side before the comparison can carry the weight placed on it.

These qualifications point to a broader way of thinking about LLMs and the RT problem that we advance in this book. The point is less about the findings in \citet{barrie2024replication} and more about \textit{how} we frame them. When applying LLMs to research, we need to understand the different sources of variation in model outputs, and then address each with a standardized set of procedures and recommendations. Within-model variation is a problem of replicability, while across-model variation is one of robustness. For within-model variation, we start from the architectural reality that decoder models are not deterministic. Aggregated over repeated iterations, however, their outputs let researchers estimate whether a robust central tendency and a constrained variance exist for a given task and model. 

This is the logic of \textit{probabilistic} RT, which we develop in Section 5.2: by using repeated sampling from a single model, we can recover a stable central tendency together with a defined measure of variation. For across-model variation, we extend the implementation guidelines set out in Section 4.11 as the BCH approach. More generally, we insist on the importance of `theory', meaning the principles that link model architectures, tasks, outputs, and downstream results. More theoretical clarity goes a long way toward helping applied researchers produce work consistent with RT principles. For us, this involves developing strong expert-coded datasets, testing more than one model, using fine-tuned models for benchmarking, and reducing measurement and other biases in downstream estimates. In the rest of the chapter, we develop these ideas and introduce our solutions to the reproducibility problem. We begin with the shift toward probabilistic RT.

\section{Thinking in Distributions: A Probabilistic Approach to RT}

At the heart of our approach is the idea that we can think of RT in probabilistic terms. LLM output is a distribution, not a fixed value. So any single run with the same model and prompt on an annotation task produces \textit{one draw} from that distribution. We normally evaluate each draw with an F1 score, but that F1 score is itself a single draw from a distribution of possible scores. Thus, we run the same task multiple times to obtain a distribution of F1 scores rather than relying on any single run. In essence, our approach embraces and leverages the added uncertainty of LLMs to produce replicable and robust work. Failing that, it provides a useful diagnostic tool for RT failures.

Probabilistic RT applies to both encoders and decoders, but it does so in slightly different ways and for different reasons. Encoders pose a narrower set of RT problems as they are, under the right conditions, essentially deterministic. With seeds and the environment fixed, the same input will always produce the same output. However, as we saw in Chapter 3, performance is highly dependent on where we draw the train-test split, particularly in small or imbalanced training sets. A favorable split will overestimate model performance and an unfavorable one will underestimate it. This is why we use 10-fold cross-validation and repeat it 10 times, which provides a distribution of scores for more robust out-of-sample estimates. The point to notice is that 10-times repeated 10-fold cross-validation \textit{is already} probabilistic RT.

Decoders, by contrast, are probabilistic at inference so they will produce different outputs across model runs even if the input is the same. The variation is thus \textit{in} the model, which complicates replicability. However, we argue that probabilistic generation also means there \textit{is} a distribution of potential outputs, one that lets us distinguish consistent from anomalous results during replication. By running the model repeatedly on a fixed version and prompt, we can extract a distribution of potential F1 scores with a clear central tendency and spread. This is similar to the output we would get with 10-times repeated 10-fold cross-validation with encoder models.\footnote{Note that the source of the variation is different for encoders and decoders. For encoder models, cross-validation evaluates the sensitivity of the model to different samples of the data, while in decoder models, the data are fixed, and the variation in output stems from the probabilistic nature of inference. This problem is aggravated when using closed-weight models, where the researcher has little control over the setup and deployment environment.} So the mechanics of encoders and decoders differ, but the principle is the same: extract a distribution of scores and report its center and spread. Note that this applies to the measurement step, not yet to the downstream bias step. 

Lower F1 run-to-run variance reflects stable aggregate performance. Yet stability can still mask diverging error structures. Lower item-label instability, where fewer observations change labels across runs, generally translates into more stable downstream betas.  However, a small number of changing labels on high-leverage observations can still move downstream estimates. This is why we should analyze the distribution of downstream betas directly rather than simply assuming it is stable based on the F1 results alone. 

Stability, however, is not accuracy. A perfectly stable measure can be systematically biased, yielding a coefficient that is consistent yet wrong across runs. This is the problem that design-based supervised learning (DSL) addresses \citep{egami2024}. Our probabilistic approach selects an ensemble of models that minimizes downstream variance and makes estimates \textit{stable}. It does not make them unbiased, however. The best-performing model often yields a more accurate measure, but not reliably so since systematic bias can still appear. Bias correction remains necessary even after choosing the model ensemble. This is why DSL complements our approach \citep{egami2024}. Probabilistic RT lowers the variance of downstream betas, and DSL then works with more stable inputs, which leads to narrower confidence intervals and fewer type-II errors, as we show in Section 5.6.

\section{Prompt Design for RT}

The prompt is another source of variation that exacerbates the reproducibility problem. Although \citet{barrie2024replication} discuss prompt stability, they do not test it explicitly in their design. Small changes in wording can shift a model's output (see Section 4.8). Researchers are neither expected nor accustomed to reporting every prompt they tried on the way to a result. This leaves the door open to both honest error and selective reporting \citep{baumann2025large}. 

In the spirit of \citet{marvin2023prompt}, we propose a standard scripted prompt that balances consistency with the elements that maximize performance. We combine this first suggestion with the rolling-window approach introduced in Section 4.8 \citep{timoneda2026rolling}. By accumulating prior labeling decisions in the context, researchers can reduce the influence of any single phrasing choice. Also in Section 4.8, we test four factors that can leak variation into the model output: prompt length, specificity, order of categories, and specified role. We model our prompt script around these elements.   

The script standardizes order and content. It includes the elements that prior research and our own testing have identified as producing greater variation in output. It is also transparent and lowers idiosyncratic content choices. At the beginning of the prompt, we suggest researchers clearly spell out the LLM's role and a context for the task. Both narrow the space the model attends to, signaling which parts of the text, semantic or syntactic, it should prioritize for the task.\footnote{Research finds that varying the persona in the prompt has little effect on outcomes \citep{zheng2023helpful}. We confirm this finding in Section 4.8. We standardize it nonetheless, to guard against unknown sources of variation in future models.} While category definitions will vary by task and researcher, we recommend giving the model hints about which linguistic signals point toward a category (positive indicators) and which point away (negative indicators). We recommend positive and negative indicators because they make the boundaries of each category explicit (what falls inside a category and what does not), particularly in cases where two categories are hard to differentiate. 

To avoid useless answers from the model, and to allow for a cleaner tidying of the data after running the code, we specify the output format. Toward the end of the prompt, we include the examples (if carrying out few-shot learning), and the space for the context window (if carrying out a rolling window approach). While there is still space for discretionary decisions regarding the wording of the task, and the definition of the concepts, we believe that following a similar script reduces sources of variation that can affect outcomes and, eventually, replicability. When replicating approaches, it also provides a format in which it is easier to identify where prompting can be tested (e.g., randomization of categories, sample of examples) and improved. 

\noindent\textbf{The standardized prompt script.} Fields in brackets are completed by the researcher; parenthetical fields are optional.
\small
\begin{verbatim}
Role (optional): You are a [role] working on [task] for [audience].

Task: You are [classifying] texts from [data description, e.g.,
Chilean congressional speeches, 1980-2020]. Assign each text to
one of [N] categories. You are identifying [concept, e.g.,
different ways of talking about poverty].

Read each passage carefully. Attend to [the features relevant to
the task, e.g., sentence structure, verb choices, and the role
poverty plays in the text]. (Few-shot / rolling window only:)
Prior classifications inform the task; stay consistent with them
unless the new text is clearly different.

Category definitions and decision rules:
[Category 1]: A text in which [definition].
  Positive indicators: [signals of membership].
  Negative indicators: [signals of non-membership].

Boundary cases (optional):
Between [Category 1] and [Category N], ask:
Is [condition that differentiates them]?
  If yes: Category 1. If no: Category N.

Output format: Respond with a JSON object only. [Further output
constraints, e.g., no explanation or additional text.]
{"label": "<category 1|...|category N>",
 "confidence": "<high|medium|low>"}

Examples (few-shot): [5-10 labeled examples per category]
Context (rolling window): [prior annotations]
\end{verbatim}
\normalsize

\section{Replicability of Encoder and Decoder LLMs: Problem and Solutions}

\subsection{10-Times Repeated Cross-Validation for Encoder Models as an Approach to Replicability}

As previously discussed, models like BERT and RoBERTa are essentially deterministic at inference. For a classification-tuned encoder, inference is a fixed sequence of arithmetic operations. The initial input embedding gets multiplied through the frozen weight matrices, layer by layer, producing a set of logits from which we select the highest one as the predicted class (argmax). Since the input text and weight matrices are the same and all we extract is one set of logits, the computations are the same each time. This contrasts, of course, with decoder models, which probabilistically output different tokens in a sequence, yielding a potentially different combination of outputs each time. For now, let us focus on encoders.

We can interpret the determinism of encoders in two ways for replicability. First, we could argue in favor of fine-tuning the model once and freezing the train-validation-test split, which  guarantees exact replication. A purist might prefer this, since it returns the same results each time. However, we think this approach has a key drawback. It relies on the observed model performance on \textit{one} random data split being representative of the \textit{true} out-of-sample model performance. As we saw in Chapter 3 and our discussion of `lucky folds', the reality is that where you split the data affects the performance you observe.

As we argued in Chapter 3, we recommend 10-times repeated 10-fold cross-validation for tasks like the ones in this book. With RoBERTa, fold-to-fold variance is small but not trivial, as one fold can score 0.81 and another 0.9. Taking any single fold as the result, even after setting a seed, risks overestimating or underestimating performance, unless one gets lucky, and those odds are not worth playing. Repeating 10-fold CV ten times yields a distribution of 100 different out-of-sample F1 scores, and its mean and standard deviation are the best approximation of true out-of-sample model performance. 

How does this bear on replicability and robustness? Well, it may complicate replicability as we depart the comfortable, deterministic world of traditional replication. But repeated resampling, averaged over randomized draws, is precisely what we mean by probabilistic replicability. If a researcher ran 10-times repeated CV on the same task using RoBERTa, the mean and standard deviation across all 100 resulting runs should be almost identical to ours. A parallel way of thinking about this is that two researchers, using the same data, model, and hyperparameter choices, should produce practically the same \textit{distribution} of F1 scores over 100 runs. To us, this is the approach that best balances the need to estimate a model's true out-of-sample performance with the requirement that results be replicable. 

\subsection{Within-Model Variation for Decoders}

Probabilistic replicability applies even more strongly to decoders, which are themselves probabilistic by nature. First, a note on scope: here we focus on within-model variation and thus \textit{replicability}, not robustness, which is the larger, more problematic issue considering the wide array of available LLMs these days. We cover robustness in the next section. Here, we think of replicability as a within-model problem because any replication of an LLM-derived result occurs within a clear and confined model space. That is, when a researcher uses an LLM to produce work, they must disclose the model version, the exact prompt, and the hyperparameters set. This focuses the replication and limits the potential for mismatches. The questions inevitably become more about whether the researcher's \textit{choices} are correct, rather than untraceable variation in the model itself. 

In practice, a decoder LLM often produces similar results across runs, especially at temperatures close to zero. We saw in our results in Section 4.6 that decoders are especially consistent across same-model runs. Gemma 4, for example, produced F1 scores of 0.5128, 0.5124 and 0.5144 for the political advertising task, and other models behave similarly. Results, then, should replicate consistently. Thus, in probabilistic replication, we are interested in the shape of the distribution, not in identical results.\footnote{For extra confidence, we can always increase the number of runs to generate distributions of means and standard deviations that converge across multiple runs.} It requires that the distribution of performance, as measured through F1 scores, dispersion, and item-level disagreement, falls within a narrow, reportable band. We need to become comfortable with small differences across replications, provided we can attribute them to model randomness rather than model choice. 

Therefore, replicability with decoders is not inherently problematic. Researcher decisions, however, can be, especially around model choice and prompting strategy. Indeed, prompt design is one of the most important decisions affecting both performance and robustness. For replicability, prompting is not an inherent issue because it is held fixed, as we reuse the author's exact prompt to replicate a full pipeline. So prompt variation is not central to replicability, but it is key for robustness, where we ask whether the result survives different prompt formulations.

\subsection{A Guideline for Replicability of Work Using LLMs}

Our approach to replicability shifts the focus from obtaining the same output, to obtaining a similar distribution. For encoder models, researchers should report the pre-processing steps, random seeds, hyperparameters, software, models used (and from where and when they were downloaded), and complete code pipeline. Researchers should also release training data, fine-tuned models and tokenizers when appropriate. When implementing our suggested 10-times repeated CV approach, the averages of performance statistics and their standard deviations must be reported. 

For decoder models, researchers should report the \textit{exact} model version, prompt, temperature and sampling settings, software, and date of access (if using an API).\footnote{We stress model exactness because most developers release many versions of the same model. For instance, OpenAI's API includes models such as GPT-5.5, GPT-5.5-pro, GPT-5.5-chat-latest, and others --- the ones with `latest' silently update.} Similarly, they should release the data and, where possible, the code. As with encoder models, researchers should run their tasks multiple times and report aggregate variation. The number of runs should be based on the degree of uncertainty observed. 

We end by reiterating that a successful probabilistic replication does not require identical labels or identical distributions. It requires the replication estimates to fall within certain pre-specified margins of the original results, for instance within two standard deviations of the reported mean performance, and comparable bands for dispersion, item-level disagreement, and downstream coefficients.

\section{Robustness of Encoder and Decoder LLMs: Problem and Solutions}

While replicability is about generating an author's exact results, robustness refers to the consistency of the findings across multiple LLMs. A study is robust if performance and downstream results hold when deploying different models. For Transformer LLMs, the question inevitably arises: if models are probabilistic in nature, many different models exist, and none is \textit{fully} transparent about how they are trained,\footnote{The only notable exception is AI2's OLMo family of models \citep{groeneveld2024olmo}.} then how can research that actively uses LLMs be robust at all? \citet{barrie2024replication}, among others, raise exactly this valid point.

We understand and share the spirit of the critique and have sought to reframe the debate slightly in a more positive direction in the introduction to this chapter. Here, we want to provide specific solutions for the core issues leveraging the benchmark-compare-harmonize (BCH) approach, which we think provides clear answers for the key issues in the general reproducibility debate. The first issue is the across-model problem, to which we turn now.

\subsection{The Across-Model Problem}

As we discussed earlier, decoder LLMs are constantly developed, introduced, deprecated and eliminated. This certainly affects replicability, as we need to be able to access models to replicate work, but it also introduces problems for robustness. The main one is that many models exist and rapidly change, so which results should we trust? Can we even trust \textit{any} results at all, in this context?

We think there is a general need for both theoretical and applied direction when selecting LLMs. We like making the parallel to statistical inference: we know that, for certain types of data structures, a certain model is more appropriate than another one. If we have survival data, well, a simple OLS regression may not suffice. Or if we have TSCS data with a binary dependent variable, OLS might work great sometimes but logistic regression may generally be more appropriate. Here, while the nexus between LLMs and data is more diffuse, we think it still matters. 

As we have argued, encoders still work quite well in annotation tasks with shorter texts and complex categories, and in topic modeling to cluster documents. Decoders, on the other hand, generally fall short on annotation tasks and need QLoRA fine-tuning to approach or improve on encoder performance. Decoders, however, complement encoders on several of these tasks. They can label and interpret the output of topic models, and they enable flexible semi-supervised variants of them (Section~4.7). For encoders, model choice is straightforward and we recommend RoBERTa, which more consistently outperforms other encoders in classification tasks in extant testing. As Chapter 3 discussed, domain-specific and multilingual encoders suit particular tasks, but RoBERTa remains the practical default. 

As for decoders, the situation is more complex. We favor open-weight models for their increased transparency and ease of fine-tuning, so that narrows the scope already. Among these, smaller models with $\leq$ 9 billion parameters tend to perform admirably well but still fall short on some tasks, having less general pattern knowledge and weaker reasoning. Small but more capable models in the 20--40B range often offer the best of both worlds, with strong performance at a reasonable cost. Chapter 4's testing confirms this with Gemma, Llama and Mistral, but performance remains task-specific. Medium and large models, 70B or above, tend to be difficult and expensive to deploy locally, and their increased reasoning capacities often do not translate into task-specific performance gains --- we saw this in Section 4.6. This, of course, is the state of the art as of this writing, and it will likely have changed by the time the book is in the reader's hands. Yet the main takeaway is the \textit{relative} standing of the models, which ultimately affects performance and choice. 

The question for robustness is how to know whether results survive model choice when we generate data with LLMs and report downstream estimates. For this, we return to Section 4.11 and its main recommendation to benchmark, compare, and harmonize. The procedure is as follows. We (1) use RoBERTa as the encoder benchmark, (2) run an ensemble of open-weight models with few-shot CoT, (3) apply QLoRA fine-tuning to the best performer for a second benchmark, (4) produce downstream estimates across models to approximate the variation in resulting betas, and (5) apply DSL to correct any remaining biases from LLM labels. Steps (1) through (3) concern measurement and how to maximize the quality of the labels, while steps (4) and (5) concern downstream inference. We approach each in turn. The following subsections develop the measurement steps: how to select the best-performing model and the ensemble (the maximization problem), and how to read the structure of errors across models, meaning when and where in the data the model errs, not just how often. We develop the downstream steps in Section 5.8.

\subsection{The Maximization Game: Selecting the Best-Performing Model for Measurement}

Measurement is a \textit{maximization} problem. As researchers, we always seek the best possible measure of a concept, and we may now want to build it from a text source using an LLM. For that, the goal is to maximize the performance of an LLM for a particular task, minimizing the measurement bias it induces. However, performance maximization is not how we traditionally approach RT, and in some ways it seems to be anathema to it. We have always focused more on the stability and `truth' of a result rather than on obtaining the best result. Yet measurement has always been a maximization exercise: the best coders, the best codebooks, and the best practices to ensure the variable that results from human-coded text is as accurate as possible. This should certainly not be different with LLMs, and we should ensure we get the most accurate annotations or topics from these models. The question, then, is how to generate the best data for downstream analysis with LLMs, just as we have long done with human coders. To us, the key to this maximization problem is to have clear guidelines for what is \textit{good} measurement with LLMs and what is not. This is what we propose here. 

First, maximization in measurement requires a gold-standard, high-quality and theoretically grounded training and/or validation set \citep{davila2026machines}. As stated throughout the book, the final measure of a concept is only as good as the definition of that concept and its identification in the data. The maximization game becomes a test of which model best executes the task, judged against the human-coded sample. Second, and as we have argued throughout, researchers should prioritize open-weight models for transparency and the ability to save and re-run exact versions \citep{spirling2023open, barrie2024replication}.

Third, we should maximize classifier performance using an \textit{ensemble} of LLMs, as the BCH approach proposes. Three points warrant further elaboration. First, the encoder benchmark is not only inexpensive and deterministic, but it also still beats state-of-the-art decoders in many applied tasks.\footnote{If a minimally fine-tuned encoder already matches or beats powerful decoder models, that is a useful signal: further gains are more likely to come from improving \textit{separability} in the codebook and training data than from switching models. At that point, therefore, the researcher can either (1) expand the labeled set to further fine-tune RoBERTa or the best-performing decoder with QLoRA; or (2) if performance is already acceptable, use the encoder to label the remaining data.} Second, the ensemble of open-weight models should span developers and sizes. At the time of writing, strong options are Gemma 4 31B, Llama 4 Maverick, and Mistral 3.2 24B, all with few-shot CoT.\footnote{Depending on computational resources and access, which we have shown are not as problematic a barrier as they once were, researchers can use the smaller 7B and 8B equivalents, if necessary.} Third, QLoRA fine-tuning of the best-performing open-weight decoder can improve performance significantly \citep{halterman2025codebook}, although gains are task-specific (see Section 4.9). Researchers with remaining funding can add the latest proprietary model from OpenAI, Anthropic or Google as a final comparison point. 

Now, we understand that this multi-model approach may increase costs, but in our experience these costs remain quite manageable when using open-weight models.\footnote{If one has access to an institutional GPU, then the researcher has little to no costs. Otherwise, costs are either from GPU rentals, which often range between 0.5 and 1.5 dollars per hour for the models and tasks tested in this book, or from the DeepInfra (or similar) API, which charges very small sums per token for open-weight models. QLoRA fine-tuning requires a GPU in any case.} There is also a clear reasoning behind it: as our results in Section 4.6 showed, performance may differ across models and tasks, and the patterns are not entirely clear.\footnote{And as we showed in Chapters 3 and 4, it is crucial that all tests are done on test data and are representative of true out-of-sample classifier performance.} One thing that \textit{is} clear: running multiple models helps us pick the best model, which is the one that maximizes out-of-sample classifier performance for a given task. As we will see later, using an ensemble also has important positive consequences for downstream confidence intervals.

An added benefit of this multi-model approach is that it allows us to get a strong overall understanding of LLM performance on our data. This in itself is valuable when reporting model choices. Currently, researchers often run the one model they know and take its performance as the best available. But that may not be the case, as a high-performing model in benchmarks may be a poor specific-task expert. Getting a holistic view of LLM performance with our data will help us confidently defend our model choice and communicate that in a transparent, convincing way. In all, selecting the best model maximizes measurement quality, but robustness depends on whether the finding survives different models and how these models differ. We turn to this key question now.

\subsection{Error Structure and Disaggregation}

By applying the BCH approach, we are in a position to select the best performer or set of performers for our task. Yet one problem remains that is key for RT: just because one model produces a better F1 score, it does not mean that it produces the best \textit{measure} of a concept. If a given model far outperforms the rest, then it most likely does, but a single model rarely dominates. Most likely, a few LLMs will yield good results, and at that point we still must decide which model's predictions we select for downstream inference. Do we select a best performer, or find a way to aggregate across them? This is, at the heart of it, one of the key issues in \citet{barrie2024replication}. We will cover the \textit{harmonization} step between LLM outputs and regression betas in Section 5.6.  

In our view, the answer lies in understanding not just the overall performance differences but \textit{how} models differ. Blind aggregation or choosing the best performer naively is a poor default. Instead, we ought to examine \textit{where} the models err, that is, their \textit{error structures}. Are the models making mistakes in the same region of the data, or in different parts of it? Erring in the same regions suggests certain observations are intrinsically hard. Any model trained on similar data will find these genuinely ambiguous cases difficult to classify correctly. Errors in the same regions have two sources. First, the overlapping text corpora that most models use for training reproduce strengths and blind spots across model families. Second, the intrinsic ambiguity of hard texts leads to error, as even human expert coders are likely to disagree on how to code them. 

If all models miss on largely the same observations, aggregating results will not yield any meaningful improvements, which makes aggregation redundant. In these cases, the best solution is to simply pick the best performer, while at the same time reporting the comparisons, the benchmarks and the model error structures in an appendix. When multiple models perform similarly, we inevitably introduce some arbitrariness into the decision.\footnote{Note that this does not preclude aggregating the results as we suggest below for cases where models err in different regions.} Yet picking the best performer and transparently reporting the process is a valid solution if error structures are very similar. Thus, the reassuring side of error-structure analysis is that it can yield principled, defensible model choices. And while this does not guarantee a similar error structure, when the procedure is repeated with different model families or versions, it makes similarity considerably more likely.

More problematic is the case where error from various models is spread across different regions of the data. This suggests that performance is linked to differences in the training of each decoder, which poses a serious problem for robustness. We can still take the highest performer, but because error structures are different, the measure will inevitably carry the biases from the chosen model and affect downstream results.  

In these cases, we argue that one solution is aggregation. When models err in different places, combining their predictions can allow the correct labels from one model to offset the mislabels from another. Say model A errs systematically on one category because of its training. By aggregating its labels with B's and C's, we can get a more reliable annotation for that specific observation. In our case, because the BCH approach uses at least five different models (RoBERTa, three few-shot open-weight decoders, and a QLoRA-fine-tuned decoder, plus optionally a closed-weight model), we can use a broad set of annotations for aggregation.\footnote{A note on RoBERTa: we include it in the final set of results for aggregation even though it is the only encoder in the list. The reason is that we care about producing the best final measure possible, and RoBERTa's labels can be as strong as a decoder's, and often better as Sections 3.7 and 4.6 show. So RoBERTa is both a benchmark in the measurement stage and part of the pool in the aggregation stage.} From there, we can think of LLM codings in the same way we think of annotations from multiple human coders: we evaluate them using inter-coder reliability metrics and ask whether the final measure should require unanimity, four votes, or three. 

One option is majority voting, which is simple, transparent, and easy to reproduce. One important advantage is that it provides information about each classification's reliability. A unanimous classification provides great certainty, while highly uncertain ones might be flagged for expert or manual review.\footnote{Researchers should prespecify the voting rule and tie-breaking rule, and show additional robustness by testing downstream estimation from alternative rules.} However, majority voting relies on strong assumptions, namely that every model deserves equal weight and that errors are independent across models. The latter can reproduce errors, rather than correct them. \citet{hovy2013learning}, for example, show that majority voting can fail when annotators differ substantially in competence. For decoder models, competence can translate into differences in training data and procedure. 

An alternative would be to produce a form of weighted aggregation. One option is to weight outputs by overall model performance. The logic here is that better-performing models produce more reliable labels and thus their outputs should carry more weight. Another option is to filter certain models out first, especially those whose overall performance is much below the rest. These outliers can be easily diagnosed and removed, much in the same way we would remove a bad and internally inconsistent human coder from a pool of coders. After filtering, we can run the aggregation on the surviving models. The key, however, is to drop only clear outliers, for one reason: in cases where models have different error structures, some of them may produce worse overall performance but better local performance in certain regions of the data. These differences are informative and valid during aggregation, yielding more accurate final labels. 

More fundamentally, the error-structure analysis is a robustness diagnostic in itself. It tells us whether models agree and thus a result is robust to model choice, or whether results are model-dependent. It helps us understand whether we can simply select the best model or must aggregate across model results. Thus, we believe the approach of analyzing model error structures is a strong solution to the across-model robustness problem.

Researchers should then compare the best individual model with the aggregated alternative on the validation set. If aggregation does not improve performance, researchers can use the best performing model. If it does, then the aggregation approach can be adopted. BCH does not require aggregation but rather a transparent and validated process, informed by the error structure, that guides the researcher in their decision.

\subsection{Error Structure Analysis on Three Datasets}

In this subsection, we use three datasets tested in Sections 3.7 and 4.6 to study the \textit{structure} of annotation error --- i.e., not how much each model errs, but \textit{where} in the data they do so. We investigate  whether different models make classification mistakes in similar or different areas of the data, which we argue will have important effects on downstream estimates. Where they do fail together, we examine whether the fault lies with the models or with the codebook. The data we use are from \citet{ballard2023dynamics}, \citet{trubowitz2021geopolitical}, and \citet{foy2025working}. The first two datasets have one underperforming category --- `uncivil' and `unclear' threats --- and thus are helpful when analyzing where models disagree on certain texts and categories. In \citeauthor{foy2025working}, on the other hand, all models perform relatively well across categories, and thus error structures should be less variable.

\begin{figure}[!t]
    \centering
    \includegraphics[width=17cm]{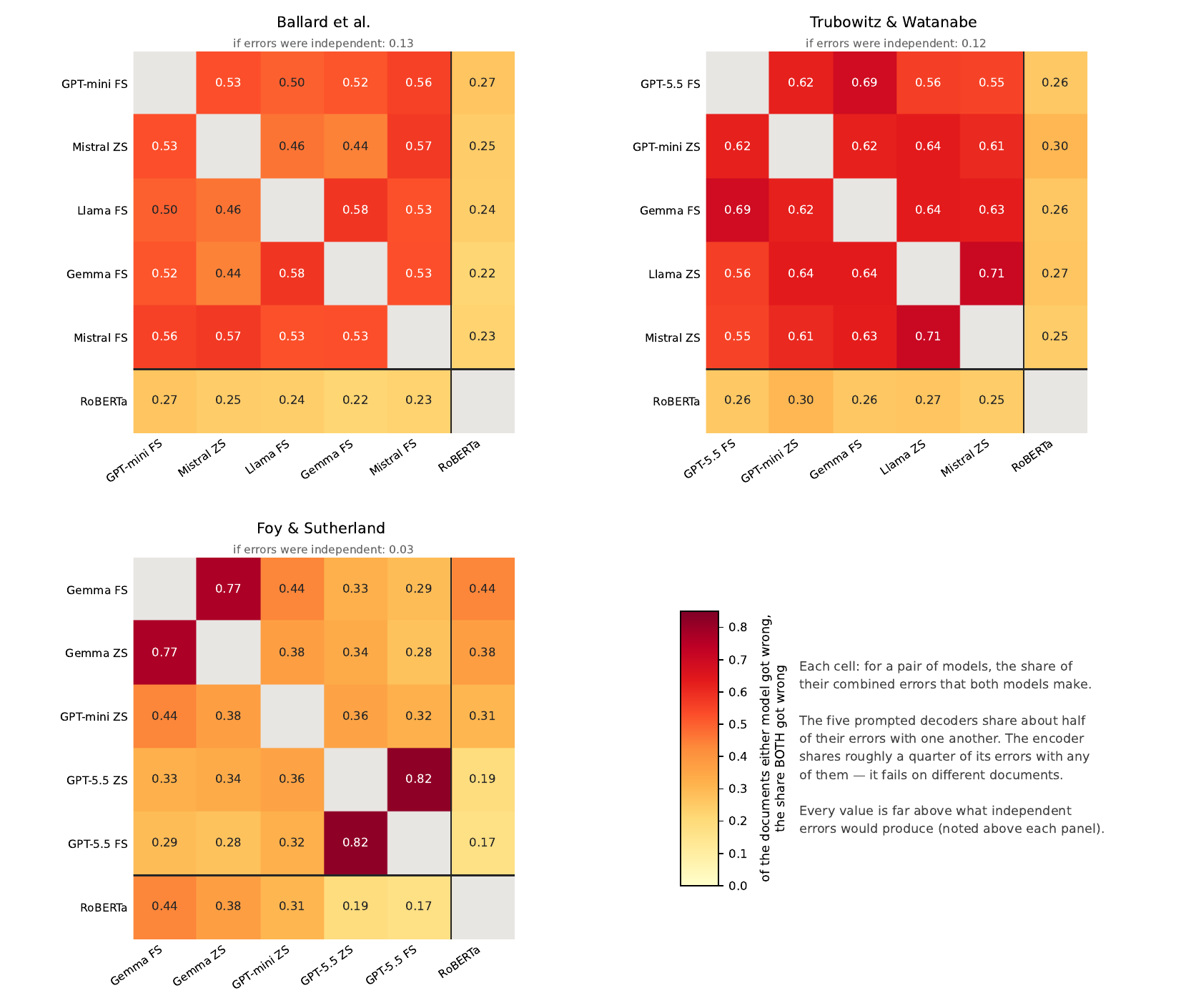}
    \caption{Error overlap across the three tasks. Each cell reports the Jaccard overlap between two models' error sets: for a pair of models, the number of documents \textit{both} models misclassified divided by the number that either one misclassified. Values closer to one indicate that the two models err on the same observations; values closer to zero indicate that they err on different observations. Lines separate the five best decoder configurations from the fine-tuned encoder. Above each panel: the overlap that independent errors would produce. In \citet{foy2025working}, the two hottest cells pair the same model under different prompts.}    \label{fig:error-overlap}
\end{figure}

First, we report pairwise error overlap in Figure \ref{fig:error-overlap}. Let $E_a$ and $E_b$ denote the sets of documents misclassified by models $a$ and $b$. Each cell represents their Jaccard overlap, $J_{ab}=|E_a\cap E_b|/|E_a \cup E_b|$: the share of observations misclassified by \textit{both} models out of all the errors \textit{either} model made. A higher value indicates that the two models tend to misclassify the same documents, while a lower value indicates that the models are misclassifying different documents. Note that the reported Jaccard overlap does not represent the accuracy of either model, or if performance would improve if we were to aggregate them; it only shows the degree of similarity of the error structure. 

The first finding is that decoders fail on the same observations more than they do with RoBERTa. In \citeauthor{ballard2023dynamics}, the mean decoder-decoder overlap is 0.52 (range 0.44--0.58), meaning that out of all the annotation errors decoders make, 52 percent are on average common to all pairs. In  \citeauthor{trubowitz2021geopolitical}, the mean decoder-decoder overlap is higher, at 0.63. These figures are far higher than independent errors would produce. If the models' mistakes were unrelated to each other, the overlap would sit near 0.13 in \citeauthor{ballard2023dynamics} and 0.12 in \citeauthor{trubowitz2021geopolitical}, the values noted above each panel. 

RoBERTa, on the other hand, produces errors that do not overlap as much with decoders --- 0.24 and 0.27 in each task, respectively. And even though those overlaps are more than twice what chance would produce, they are substantially lower than the decoder-decoder overlaps. The overall pattern would suggest that some documents are simply hard for all models to label correctly, yet RoBERTa produces a different set of predictions. In these two tasks, decoders overlap around 2.2 times more with each other than with RoBERTa. The consequences are visible at the level of the corpus. In \citeauthor{ballard2023dynamics}, nearly half of all tweets (1,332, or 46.1 percent) are misclassified by at least one of the six models. Thus, reporting only model performance would be misleading, as per-model accuracy considerably overstates how much of this data models label \textit{reliably}, and suggesting a high degree of model dependence. More telling is that 298 tweets (10.3 percent) defeat all five decoders at once, with RoBERTa recovering 182 of them.\footnote{Figure \ref{fig:error-overlap} shows that decoders share similar conceptual boundaries, probably because they are trained using similar data and procedures, and thus make similar errors. RoBERTa, on the other hand, appears to impose different category boundaries much less based on priors and more on the training data itself. Hence the importance, we think, of continuing to include RoBERTa in the loop of LLM-annotation and analysis.} Note that an aggregation rule applied to unlabeled data would still need to determine when the encoder should override decoders. Knowing that the encoder errs on different observations than the decoders does not by itself change the rule, but it shows the importance of a heterogeneous pool of models. More broadly, the results show that models producing different patterns (e.g., encoders) are valuable for the comparison and aggregation stages of BCH.  

\begin{figure}[!t]
    \centering
    \includegraphics[width=14cm]{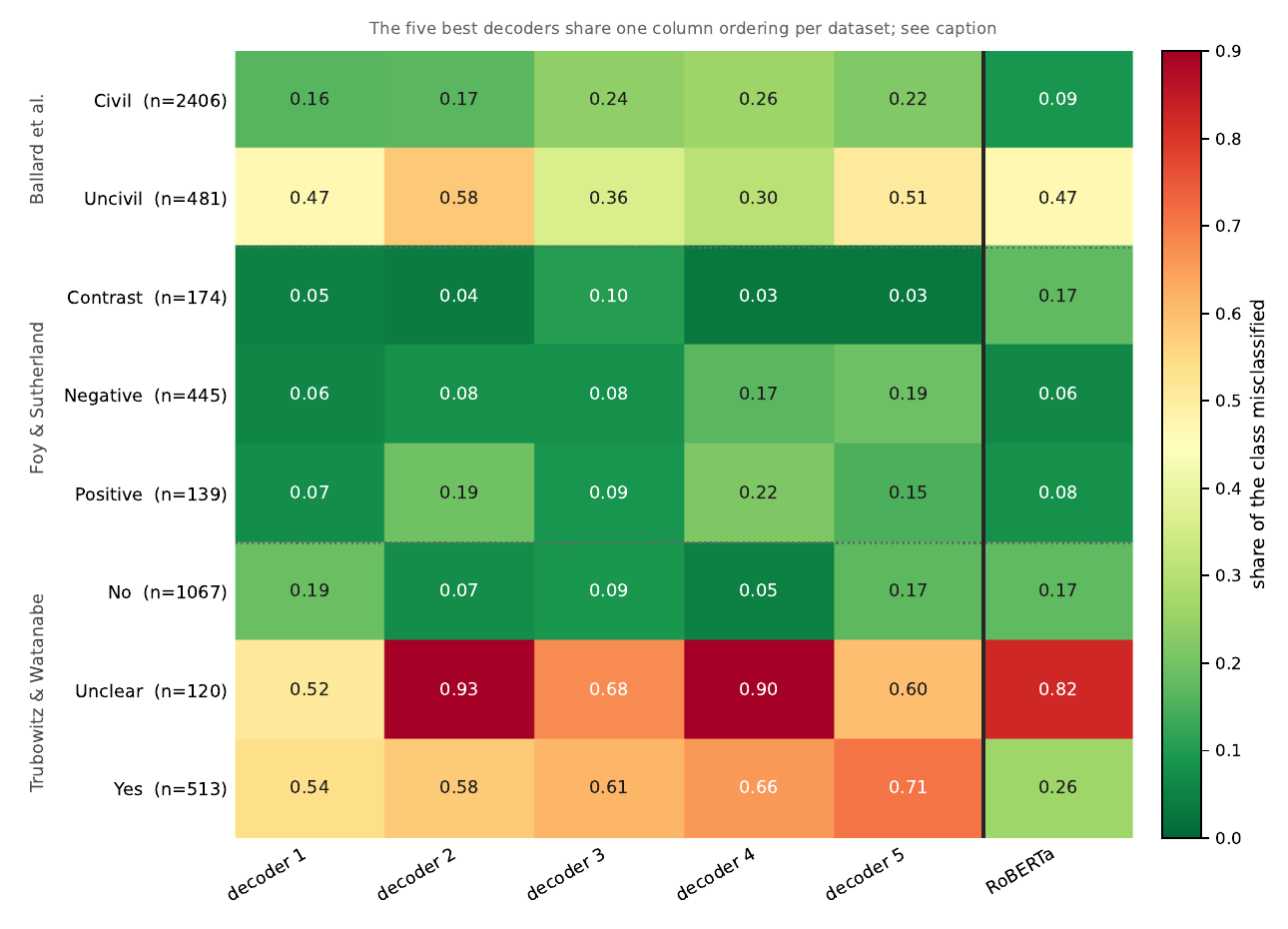}
    \caption{\small Per-class miss rates for the five best decoder configurations and the fine-tuned encoder, by dataset. Each cell is the share of the class the model misclassifies; the vertical line separates the decoders from RoBERTa. Because each dataset ranks its own five best decoders, the decoder columns denote different models in each block. In \citet{ballard2023dynamics}: GPT-mini FS, Mistral ZS, Llama FS, Gemma FS, Mistral FS. In \citet{foy2025working}: Gemma FS, Gemma ZS, GPT-mini ZS, GPT-5.5 ZS, GPT-5.5 FS. In \citet{trubowitz2021geopolitical}: GPT-5.5 FS, GPT-mini ZS, Gemma FS, Llama ZS, Mistral ZS. The encoder is more accurate than the decoders on the common classes but less accurate on several minority classes --- \textit{Contrast} in \citeauthor{foy2025working} and \textit{Unclear} in \citeauthor{trubowitz2021geopolitical}.}
\label{fig:per_class_error}
\end{figure}

Figure \ref{fig:per_class_error} breaks down errors by category and model type. On the x-axis are the models, with the top-5 decoders per task plus RoBERTa (see caption listing all decoders per task). Each cell represents the share of misclassified texts per category when compared to the human-labeled validation set. These are the class-specific miss rates (rather than overall accuracy). 

First, RoBERTa does considerably better in two cases: \citeauthor{ballard2023dynamics}'s `civil' category and \citeauthor{trubowitz2021geopolitical}'s `yes' category. It generally shows strong performance compared to all decoders, except for \citeauthor{trubowitz2021geopolitical}'s `unclear' category, where RoBERTa misses 82 percent of observations, compared to 52 percent from the best decoder. On the `yes' category, on the other hand, RoBERTa only misses 26 percent of classifications compared to the best decoder at 54 percent. 

The poor performance on the `unclear' category may reflect several factors. The category is relatively small ($n = 120$), which gives encoder models fewer examples during fine-tuning \citep[on remedies for sparse categories, see][]{timoneda2026llms}. This might explain part of the results, but RoBERTa does well in \citeauthor{foy2025working}'s positive class, also underrepresented. Alternatively, `unclear' is a category with blurry boundaries that make it difficult to consistently identify in text. On the `yes' category, on the other hand, RoBERTa only misses 26 percent of classifications compared to the best decoder at 54 percent. Decoders, across the board, confound `unclear' and `yes', and some significantly more than others. Again, the shared difficulty on these categories across models shows the importance of the codebook, which often signals the task's performance ceiling, and categories that might require additional (human) review or additional (human) annotations. 

Taken together, Figure \ref{fig:per_class_error} confirms that encoders and decoders fail on different categories, so picking only one model to build measures for downstream inference is difficult to justify. The fact that model performance changes significantly by task complicates this further. While RoBERTa often outperforms decoder models, the error structure is variable, sometimes attributed to data imbalance, sometimes to training methods and data. Model selection based solely on overall accuracy metrics can mislead and hide per-category limitations that might be substantively important (e.g., rare categories or theoretically interesting categories).

The results from \citeauthor{ballard2023dynamics} provide a good illustration of this point. The best performing decoder (GPT-5.5-mini FS) has a greater miss rate imbalance between civil (0.16) and uncivil (0.47) tweets, while the fourth best decoder (Gemma 4 FS) does meaningfully better with uncivil tweets but this comes at the expense of worse performance on civil tweets. Thus, decoder 1 and decoder 4 for the \citeauthor{ballard2023dynamics} task will produce very different measures of the same concept. These differences do not mean that selecting only one model is never the right option. A model that performs best overall, produces desired class-specific accuracy, and outperforms the ensemble after the aggregation rule, will be a better choice than the ensemble. Rather, the results show why multi-model validation is important when applying LLMs in our research pipeline.  

\begin{figure}[!b]
    \centering
    \includegraphics[width=14cm]{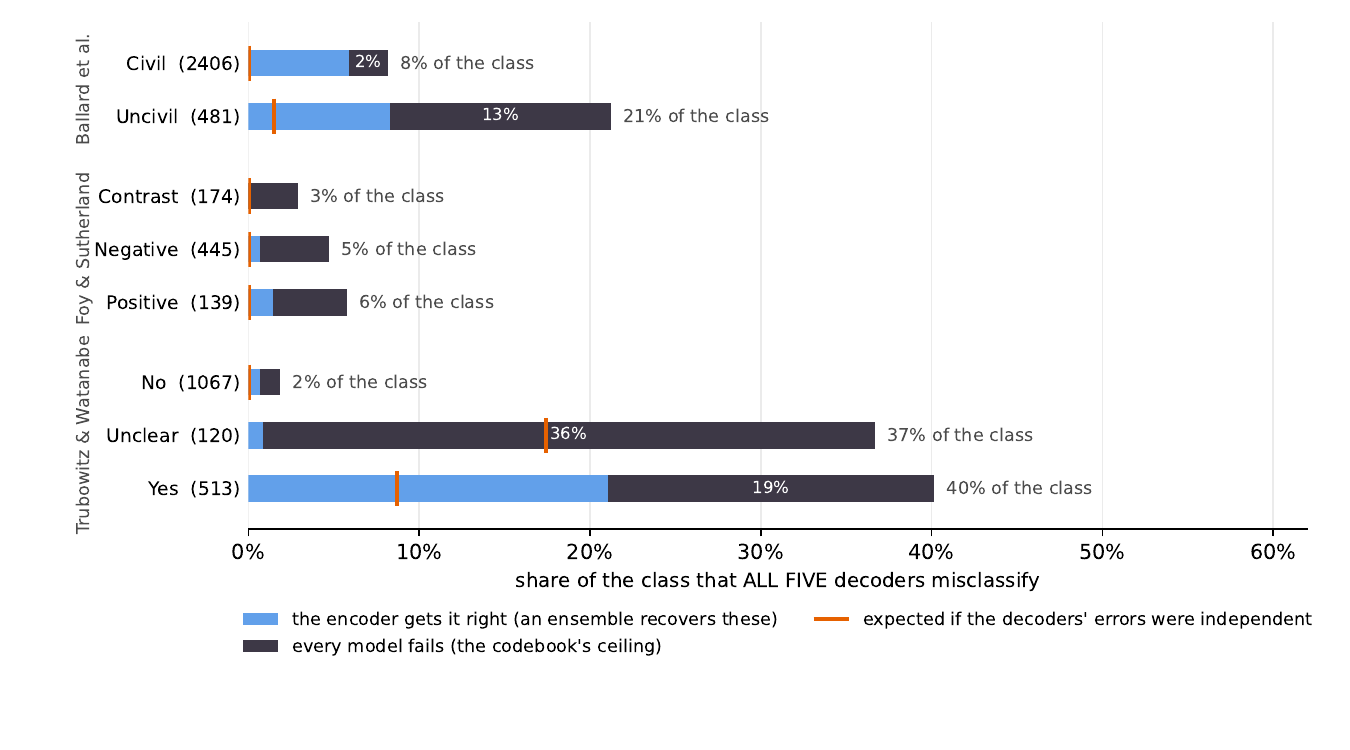}
    \caption{\small Shared decoder failures, and how many the encoder recovers. Each bar reports the share of a gold class that all five best decoders misclassify; the light portion is recovered by the fine-tuned encoder, the dark portion is missed by every model. The line on each bar marks the share that would be jointly misclassified if the decoders' errors were independent --- the gap between rule and bar measures how far decoder failures are correlated rather than coincidental. Class sizes in parentheses. On \citet{trubowitz2021geopolitical}'s \textit{Yes} class the encoder recovers over half of the joint failures; on \textit{Unclear} it recovers one document in forty-four, the codebook's ceiling.}    
    \label{fig:rescue}
\end{figure}

To conclude this section, Figure \ref{fig:rescue} shows the share of texts RoBERTa classifies correctly out of the ones that decoders get wrong. It also shows how much of the error would be expected if decoders were uncorrelated in how they classify text (vertical red line). The figure is a post hoc analysis that allows us to observe the information the encoder contributes, and not the performance of the aggregation rule. 

The results are striking. In \citeauthor{ballard2023dynamics}, all five decoders misclassify 8 percent of `civil' and 21 percent of `uncivil' tweets. RoBERTa correctly classifies a significant proportion of those misses, indicating that its errors are different from the decoders' errors. The gains are smaller in \citeauthor{foy2025working}, where all models already perform relatively well. The strongest example is the `yes' category in \citeauthor{trubowitz2021geopolitical}. While all five decoders jointly misclassify 40 percent of these texts, RoBERTa correctly classifies more than half of the observations missed. Note that selecting RoBERTa alone would not produce a lower miss rate. RoBERTa's miss rate for the `yes' category is 26 percent, and it might err on observations the decoders classify correctly. A six-model majority rule rule would not necessarily correctly identify these cases either, since five similarly incorrect decoders would still outweigh the encoder's correct vote. However, the results establish that there is a potential for improvement from the proposed ensemble approach.  

The \textit{unclear} category presents the opposite pattern. RoBERTa mislabels a similar share as the decoders. In practice, without gold labels, it is impossible to know from agreement alone whether lower performance is due to a codebook problem or an LLM blind spot for a certain category. It does point, however, to particular texts in the data that would benefit from manual adjudication, revision, or clarification. The main takeaway from Figure \ref{fig:rescue} is, therefore, that an ensemble approach where RoBERTa still has an important role can help us address the core issues of decoder performance and its differences across models.  

In this section, we have provided a series of robustness steps for LLM research. We encourage researchers to measure the performance of a heterogeneous group of models against an externally validated set, and compare it to performance after the aggregation rule, paying particular attention to the item- and class-specific error structure. Error diversity shows the importance of carrying out these tests, but it does not guarantee an improved measurement. This is still dependent on the correlation of the error structure of each individual model and the rules used to combine them. 

We have discussed simple voting rules as a transparent baseline for aggregation. They are not, however, the optimal aggregation strategy for every task. We encourage researchers to explore other rules and validate whichever one, from majority voting to performance-weighted stacking, best fits their data. The broader BCH recommendation then is that model selection or aggregation needs to be theoretically justified and empirically validated. Now, while improved measurement yields more accurate variables, we still must analyze the effects of LLM-derived measures on downstream analyses. The next section addresses this key issue.

\section{Harmonizing LLM Measures and Downstream Estimates}

\subsection{The Downstream Robustness Problem}

The challenge, as the previous section makes clear, is to produce valid, stable inference from LLM-derived measures. Doing so is difficult due to the variation both within and across different LLMs. Even small variations in the data can affect downstream estimates, especially in smaller samples. In terms of robustness, such variation signals model dependence, and questions the validity of findings built on a single model's annotations. 

Variation in LLM classification output is a well-known phenomenon, and solutions like design-based supervised learning (DSL) have been put forward to address it \citep{egami2024}. As we show in the next subsection, DSL corrects estimates toward the expert-coded benchmark, no matter how biased their coefficients are when compared to a gold-label regression. The central claim of DSL is that efficiently combining expert-annotated samples can produce valid statistical estimates, even when predicted variables contain non-random prediction errors, as is typical of LLM output. Thus, DSL can be applied as one of the \textit{harmonizing} steps of BCH between LLM measures and downstream estimators. Since DSL allows for multiple LLM outputs to enter the prediction model, we examine empirically whether an ensemble approach reduces sensitivity to any one model, and how it improves the efficiency of the corrected downstream estimates.  

Yet relying on DSL for everything is problematic, for two reasons. The first is more conceptual. The fact that we \textit{always} need a human sample to correct for LLM-induced coefficient bias \textit{and} to re-estimate confidence intervals calls into question the very relevance of the LLMs themselves. As our tests below confirm, we have no choice but to apply DSL to every task because most LLM measures induce downstream bias. The second reason is more practical. While DSL corrects for coefficient biases, it must recalculate confidence intervals because the direction and size of the error are unknown ex ante. The CIs depend on the size of the gold sample, which is often small. So despite a large sample of LLM-labeled texts, we end up with wide confidence intervals after the DSL step, leading to more type-II errors. Cross-model variation and the CI problem motivate us to develop an \textit{ensemble} approach to downstream analysis: drawing from multiple models leads to more stable coefficients and narrower CIs that remain valid.\footnote{The narrowing comes from a more informative surrogate, not from understating uncertainty; coverage is guaranteed by the DSL step throughout, so the gain is precision without a rise in false positives.} The ensemble, therefore, completes the \textit{harmonize} step between LLM measures and downstream estimators.

\subsection{Testing Measurement-Choice Sensitivity on Downstream Betas}

Our conceptual contribution concerns the \textit{variance} of downstream betas conditional on the choice of LLM for measurement. \citet{barrie2024replication} show that LLM choice can move downstream estimates, and we build on this insight by connecting it to the models' error structures and testing what the DSL correction can and cannot repair. In this section, we test our argument empirically, with two explicit goals. The first is to estimate the variance of downstream betas produced by different LLM measures. The second is to estimate whether lower variance from high-performing models yields more stable DSL corrections. For the tests, we again use data from \citet{ballard2023dynamics}, \citet{trubowitz2021geopolitical}, and \citet{foy2025working}, as in Section 5.5. The first two are high-sensitivity cases, since per-category performance varies substantially across configurations in both datasets (see Table 4.3 of the book). \citeauthor{foy2025working}, on the other hand, serves as a low-sensitivity comparison that remains stable across all decoder and encoder specifications. With this third dataset, and if our account is right, the downstream estimates should be similar across LLM measures.

The design of the tests is as follows. We hold a downstream regression specification fixed and vary only the source of the measured variable. We run separate regressions using (1) only the human gold labels, which serve as our expert-coded benchmark $\hat{\beta}_{gold}$; (2) labels from each of the separate decoder models and configurations (few-shot/zero-shot) evaluated in Chapter 4; and (3) the RoBERTa-large measures from Chapter 3.\footnote{To draw actual per-observation predictions from RoBERTa-large, we ran a single seeded ten-fold pass per dataset. The out-of-fold performance closely matches the scores reported in Chapter 3. For instance, \citeauthor{ballard2023dynamics}'s run returned a 0.722 macro-F1 compared to Chapter 3's 0.720.} Because every configuration produces its own version of the measure, it also yields its own downstream estimate of the coefficient of interest. We compare the LLM-derived betas with the human-labeled ground truth coefficients, computing the deviations and the variance across model betas. Deviations are calculated as the difference between each model estimate $\hat{\beta}_{m}$ and $\hat{\beta}_{gold}$. Lastly, while DSL's validity guarantee holds regardless of which measure it corrects, its \textit{variance} depends on how informative that measure is. More accurate predictions improve DSL efficiency, but accuracy cannot be the sole criterion for model choice. Errors correlated with the variables in the model, such as party, year, or incumbency, may matter more than other error types, even in less accurate models. Still, better precision at the measurement stage generally yields more precise DSL-corrected betas. 

We run a standard multivariate linear probability model for downstream analyses.\footnote{We keep the model choice deliberately simple and consistent across datasets to make the tests parsimonious and comparable. Whether the effects of LLM measures on downstream betas vary with the choice of statistical model is beyond our scope and should be the object of further research.} We use the LLM-derived measure as the dependent variable and include three variables from the original data sources in the regression as predictors. In \citeauthor{ballard2023dynamics}, for example, we regress the incivility label on the member's party, chamber, and gender, with the party coefficient as the main estimate of interest --- i.e., whether one party's rhetoric registers as more uncivil than another's. A final clarification is that the human-labeled data are used twice in the design. The first is to produce the ground-truth $\hat{\beta}_{gold}$. The second is to implement DSL. We simulate a researcher who hand-labels only 10 or 20 percent of the corpus and applies the DSL correction to the downstream estimates.\footnote{We also report results using 50 and 100 percent of the sample for comparison.} 

\subsection{Results: \citet{ballard2023dynamics}}

\begin{figure}[!h]
\centering
\begin{subfigure}[t]{0.545\textwidth}
    \centering
    \caption{\footnotesize{10\% gold ($n_{gold}=289$)}}
    \includegraphics[width=\linewidth]{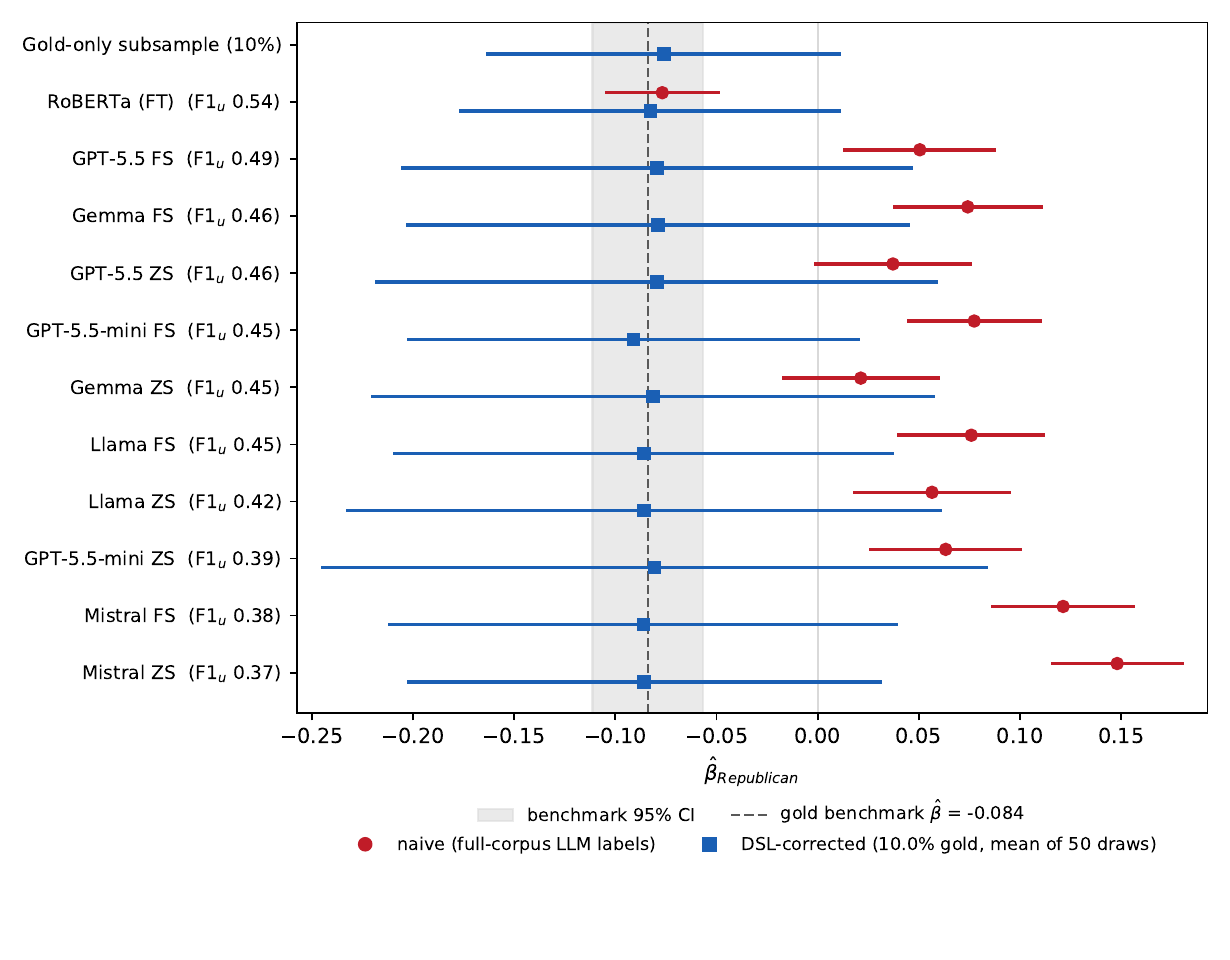}
    \label{fig:dsl-10}
\end{subfigure}\hfill
\begin{subfigure}[t]{0.435\textwidth}
    \centering
    \caption{\footnotesize{20\% gold ($n_{gold}=577$)}}
    \includegraphics[width=\linewidth]{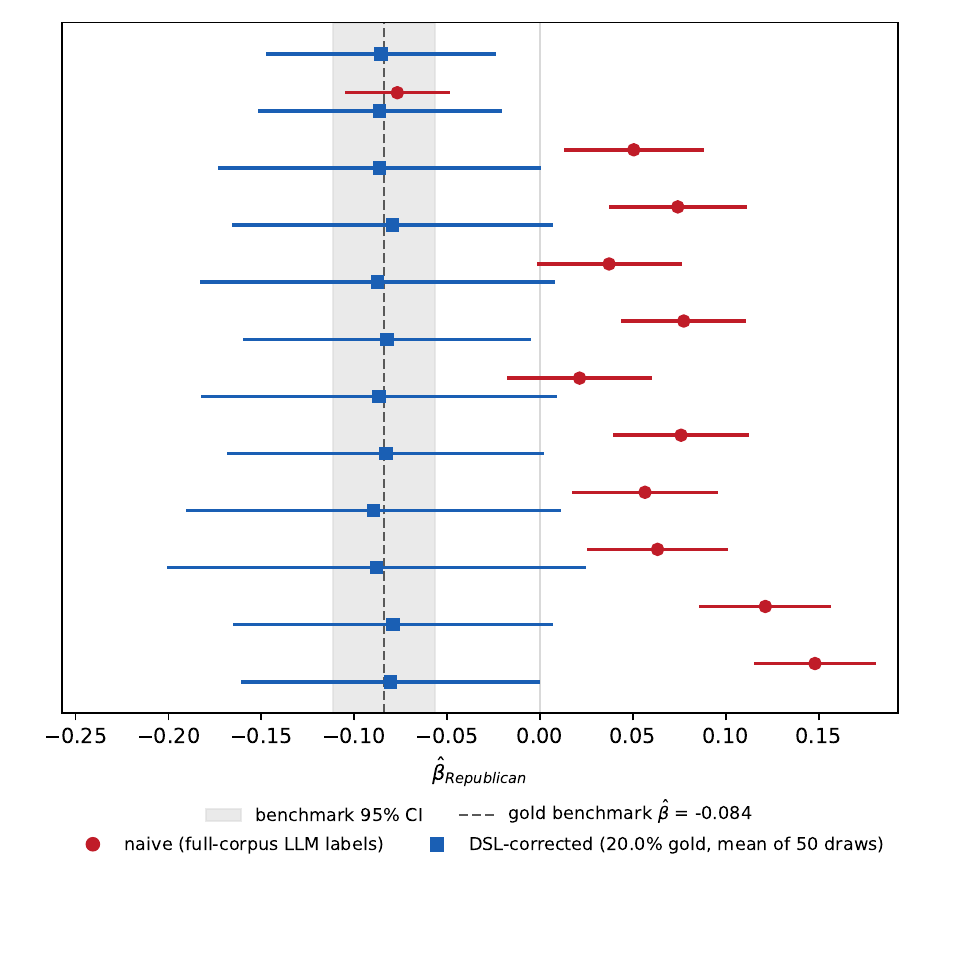}
    \label{fig:dsl-20}
\end{subfigure}

\vspace{-1.3cm}

\begin{subfigure}[t]{0.545\textwidth}
    \centering
    \caption{\footnotesize{50\% gold ($n_{gold}=1{,}444$)}}
    \includegraphics[width=\linewidth]{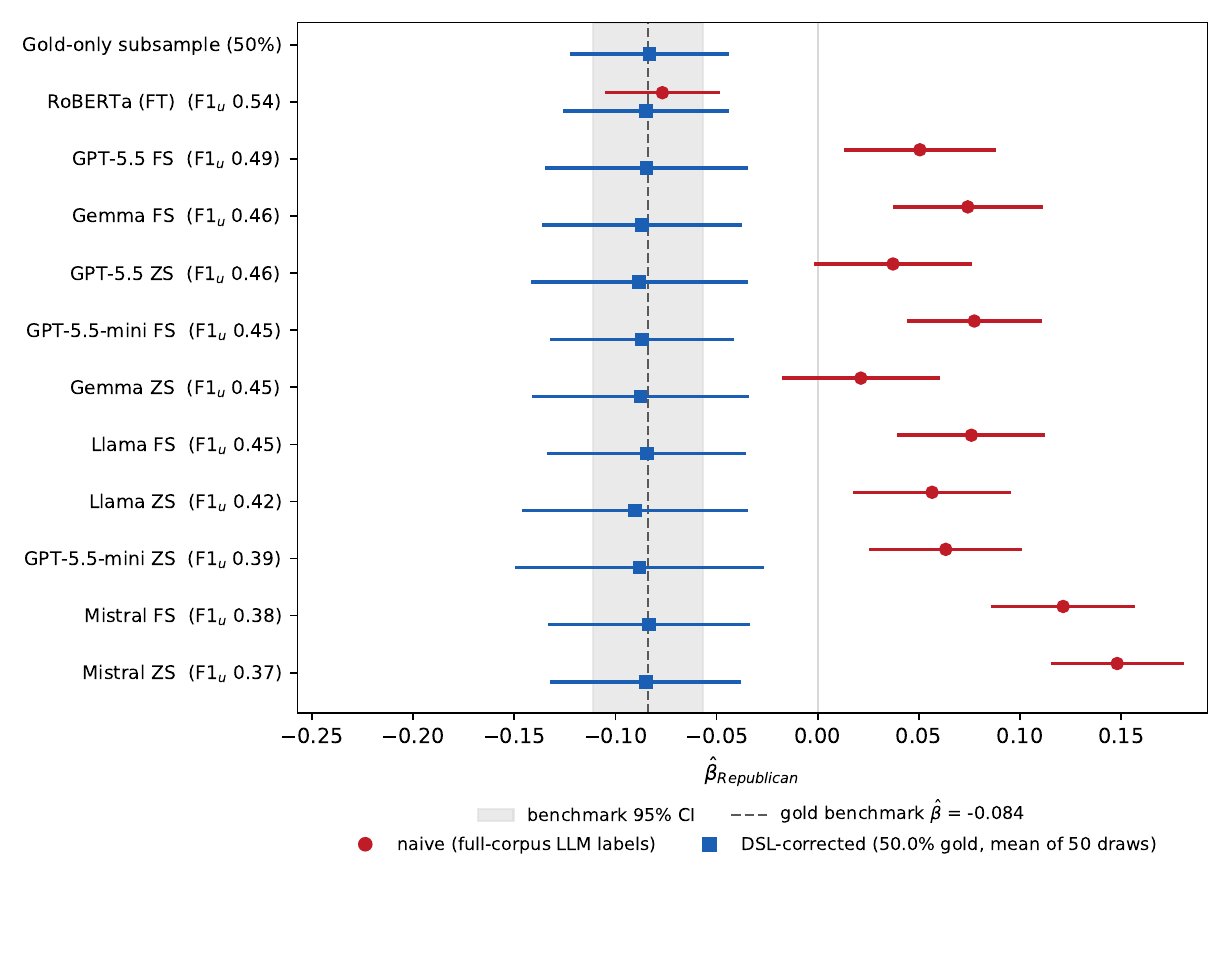}
    \label{fig:dsl-50}
\end{subfigure}\hfill
\begin{subfigure}[t]{0.435\textwidth}
    \centering
     \caption{\footnotesize{100\% gold (DSL = gold regression)}}
   \includegraphics[width=\linewidth]{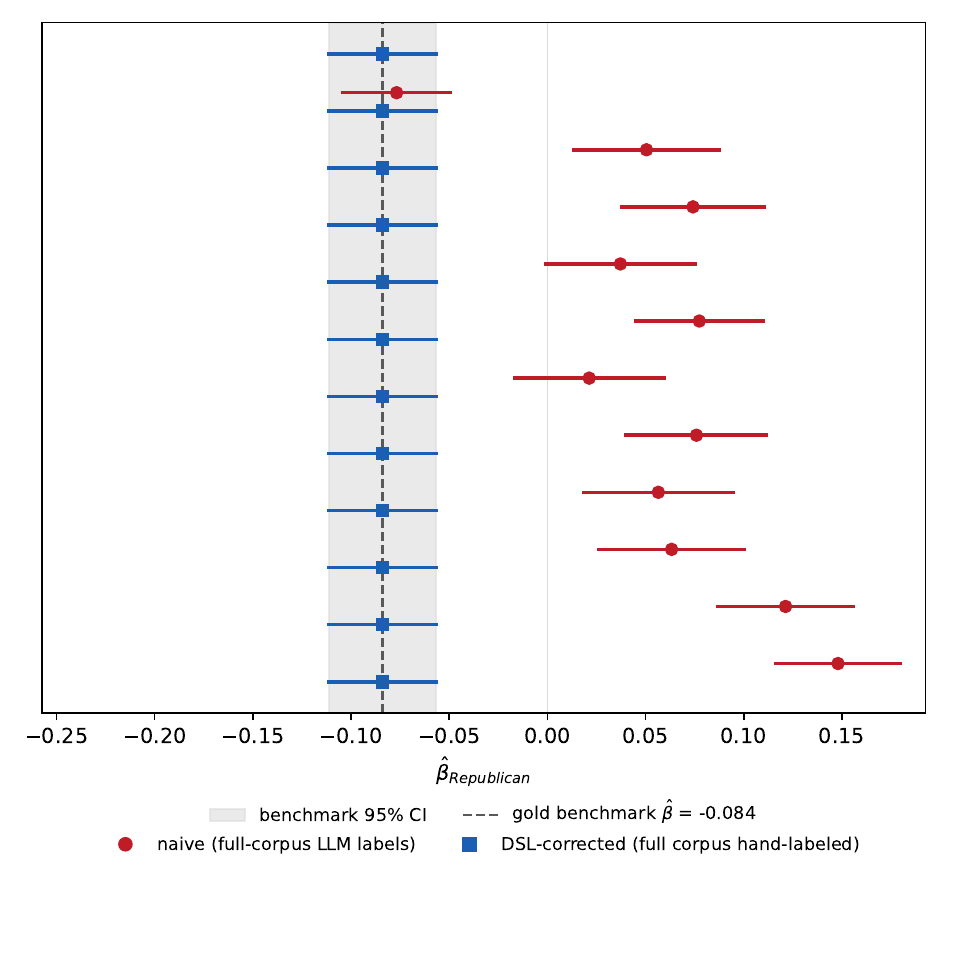}
    \label{fig:dsl-100}
\end{subfigure}
\vspace{-1.3cm}
\caption{Naive (red) versus DSL-corrected (blue) estimates of $\beta_{Republican}$ in the \citet{ballard2023dynamics} data, by hand-labeling budget. Dashed line and shaded band: gold-label benchmark and its 95\% CI. Configurations ordered by uncivil-class F1.}
\label{fig:dsl-2x2}
\end{figure}

The first set of results we introduce are from \citet{ballard2023dynamics}. We run a set of OLS regressions with the binary incivility indicator as the dependent variable and the legislator's party, chamber, and gender as the independent variables. The main coefficient of interest is  party, an indicator for Republican legislators, so that $\hat{\beta}_{Republican}$ captures the difference in measured incivility between the two parties' tweets. For efficiency, we report the full set of results for all models in Figure \ref{fig:dsl-2x2}. The dashed vertical line displays the $\hat{\beta}_{gold}$, the red lines report all model results using the LLM-derived data naively, and the blue lines show the DSL-corrected estimates. The four subplots show the results varying the level of true gold labels used for the DSL-correction step.

The results from Figure \ref{fig:dsl-2x2} are striking. \textit{All} naive measures from decoder models produce coefficients with \textit{positive} signs, and eight of ten are statistically significant, while the gold benchmark is \textit{negative}. This sign reversal is highly problematic on its own and highlights the need for correction. DSL's estimates mirror the benchmark across all models. However, despite the coefficient correction, DSL is less precise because it is estimated in a smaller sample of gold labels. Only at around 50 percent of the full sample do DSL confidence intervals begin to approximate those of the true model. In this application, RoBERTa, the best-performing model in Table 4.2 of the book, yields unbiased coefficients that mirror the results of the true benchmark without the need for correction. It also offers more precise confidence intervals almost as narrow as the gold benchmark's.\footnote{We know RoBERTa is close to the benchmark only because we hold the full set of gold labels. A researcher with a 10 or 20 percent sample can only see that their model performs well on the validation set. As Section 4.6 showed, this does not guarantee an unbiased downstream coefficient. The DSL step converts that uncertainty into a guarantee, but does so at the price in precision we document in this section.}

\begin{figure}[!h]
\centering
\begin{subfigure}[t]{0.545\textwidth}
    \centering
    \caption{10\% gold ($n_{gold}=170$)}
    \label{fig:trub-dsl-10}
    \includegraphics[width=\linewidth]{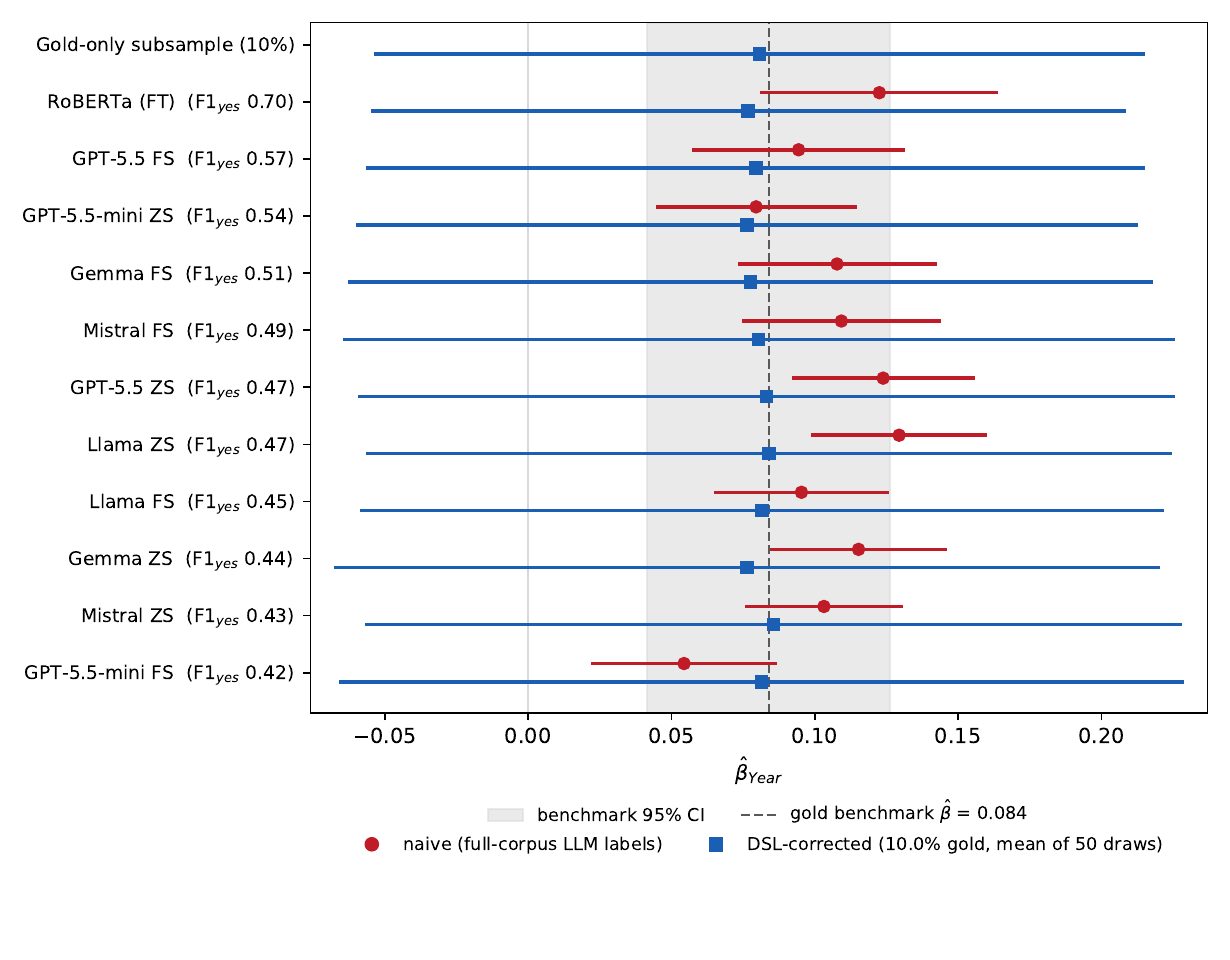}
\end{subfigure}\hfill
\begin{subfigure}[t]{0.435\textwidth}
    \centering
    \caption{20\% gold ($n_{gold}=340$)}
    \label{fig:trub-dsl-20}
    \includegraphics[width=\linewidth]{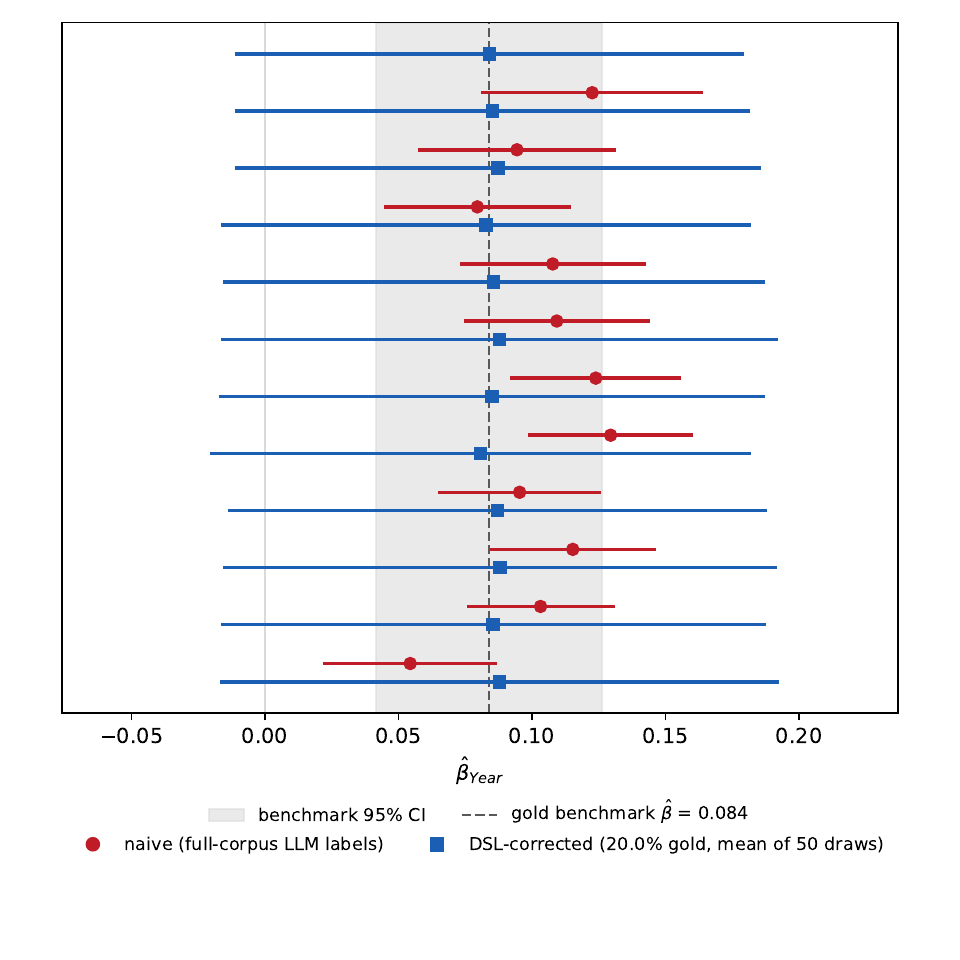}
\end{subfigure}

\vspace{-1.3em}

\begin{subfigure}[t]{0.545\textwidth}
    \centering
    \caption{50\% gold ($n_{gold}=850$)}
    \label{fig:trub-dsl-50}
    \includegraphics[width=\linewidth]{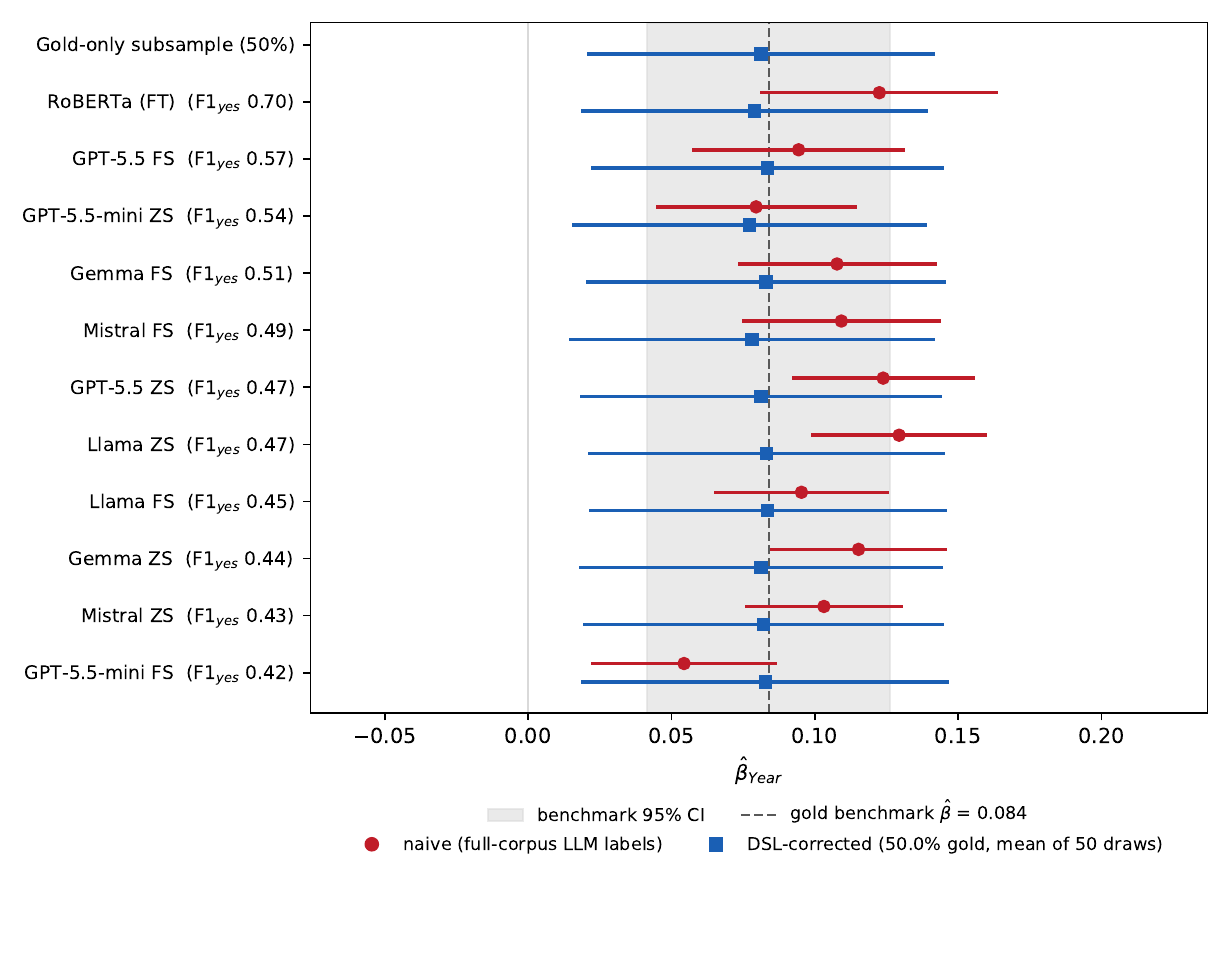}
\end{subfigure}\hfill
\begin{subfigure}[t]{0.435\textwidth}
    \centering
    \caption{100\% gold (DSL = gold regression)}
    \label{fig:trub-dsl-100}
    \includegraphics[width=\linewidth]{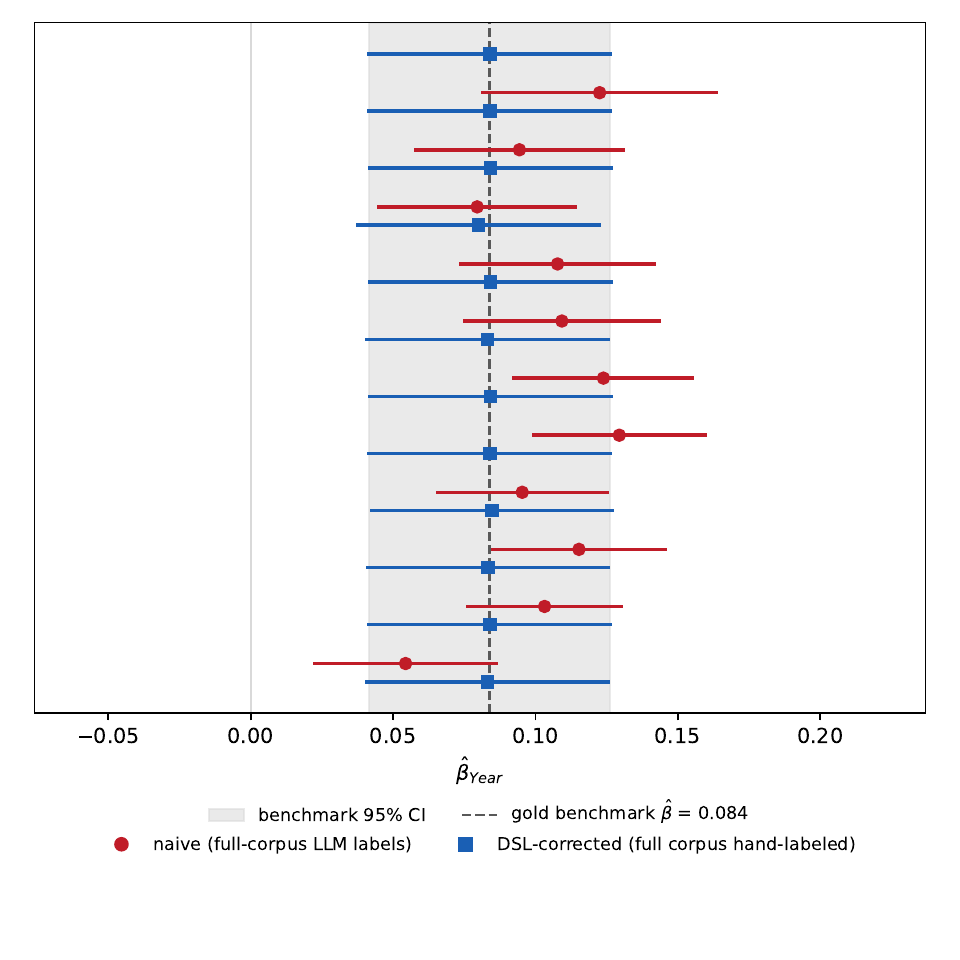}
\end{subfigure}
\vspace{-1.3em}
\caption{Naive (red) versus DSL-corrected (blue) estimates of $\beta_{Year}$ (threat-share change per century) in the \citet{trubowitz2021geopolitical} data, by hand-labeling budget. Dashed line and shaded band: gold-label benchmark and its 95\% CI. Configurations ordered by yes-class F1.}
\label{fig:trub-dsl-2x2}
\end{figure}

\subsection{Results: \citet{trubowitz2021geopolitical}}

Figure \ref{fig:trub-dsl-2x2} reports the results for the \citeauthor{trubowitz2021geopolitical} task using the same structure as the previous figure. The dependent variable is dichotomous and captures whether an article conveys a political threat. The main explanatory variable is year, and the controls are `foreign' and `wartime'.\footnote{We take `year' as the focal predictor because it represents an intermediate case: model errors correlate with it less strongly than with `wartime' but more than with `foreign', making it a fair test of the correction.} The best-performing model for the task from Chapter 4, GPT-5.5 with few-shot CoT, performs well, as its coefficient and CIs closely mirror the gold-label benchmark. The substantive results would not change using RoBERTa, but its point estimate is still slightly higher than the benchmark, despite being within its confidence interval. There is also more variability across other decoders, with some like Mistral or Llama overstating the effect and GPT-5.5-mini few-shot understating it. As with \citeauthor{ballard2023dynamics}, applying DSL again corrects the coefficient deviations well, but the added uncertainty could affect significance readings. Ideally, in this task, we would use GPT-5.5's codings without correction, but researchers do not know the ground truth ex ante if they do not have the full set of coded samples in hand.

\begin{figure}[!ht]
\centering
\begin{subfigure}[t]{0.545\textwidth}
    \centering
    \caption{10\% gold ($n_{gold}=76$)}
    \label{fig:foy-dsl-10}
    \includegraphics[width=\linewidth]{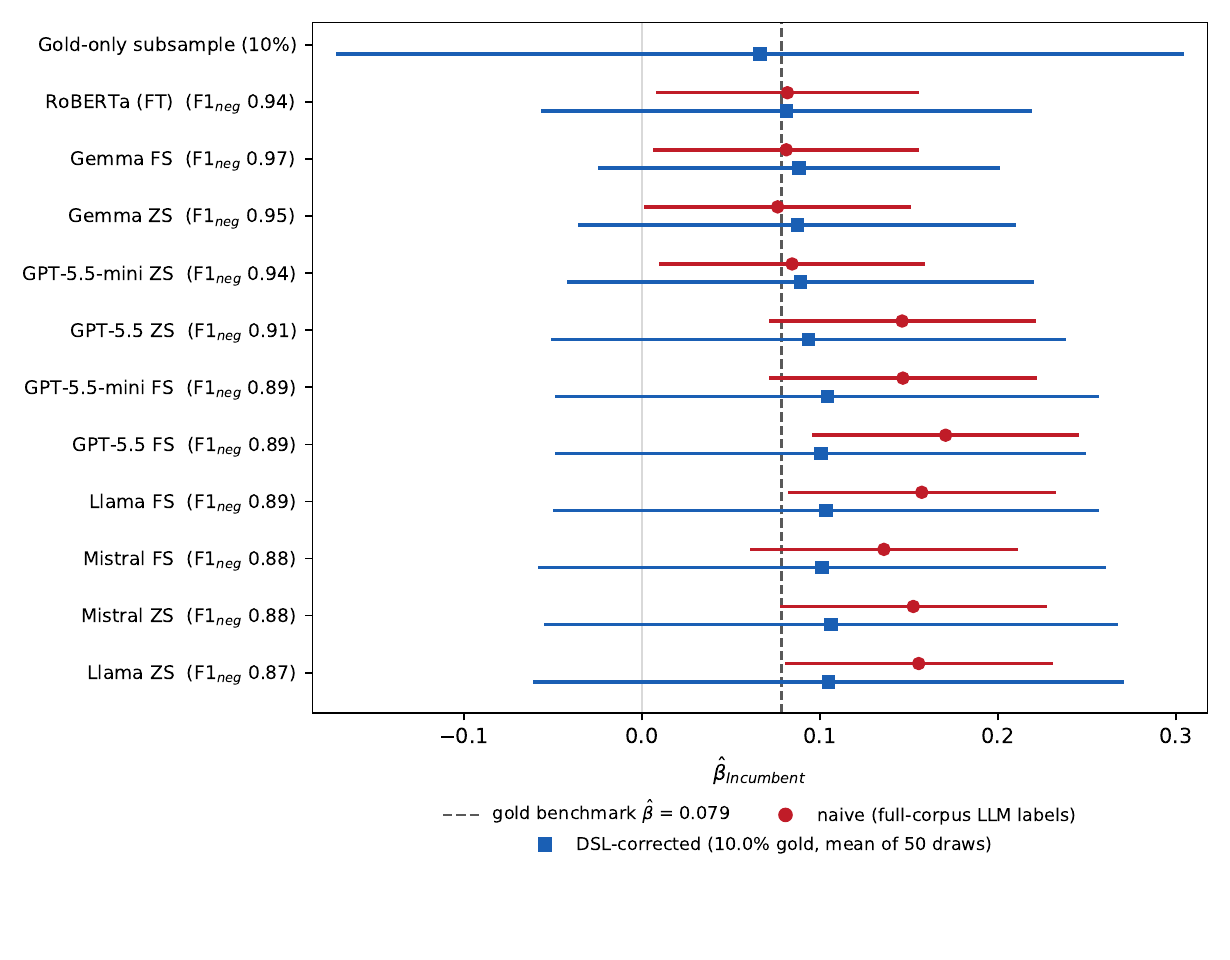}
\end{subfigure}\hfill
\begin{subfigure}[t]{0.435\textwidth}
    \centering
    \caption{20\% gold ($n_{gold}=152$)}
    \label{fig:foy-dsl-20}
    \includegraphics[width=\linewidth]{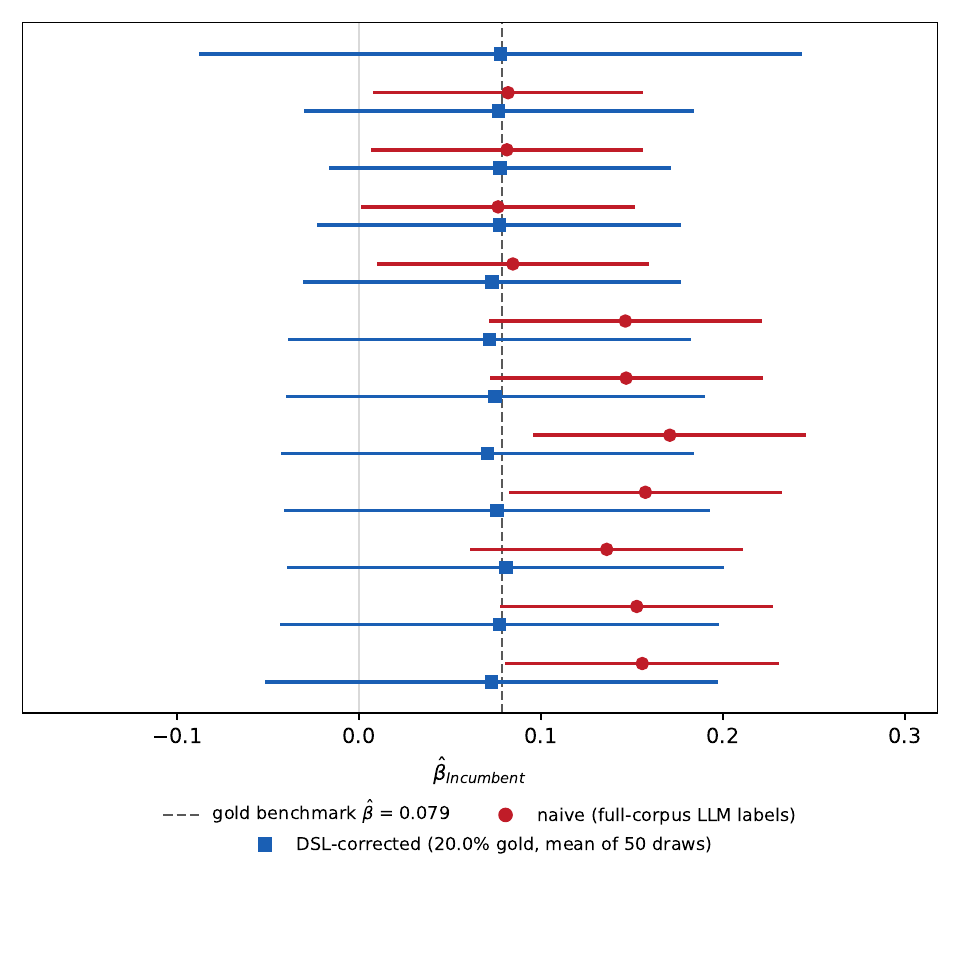}
\end{subfigure}

\vspace{-1.3em}

\begin{subfigure}[t]{0.545\textwidth}
    \centering
    \caption{50\% gold ($n_{gold}=379$)}
    \label{fig:foy-dsl-50}
    \includegraphics[width=\linewidth]{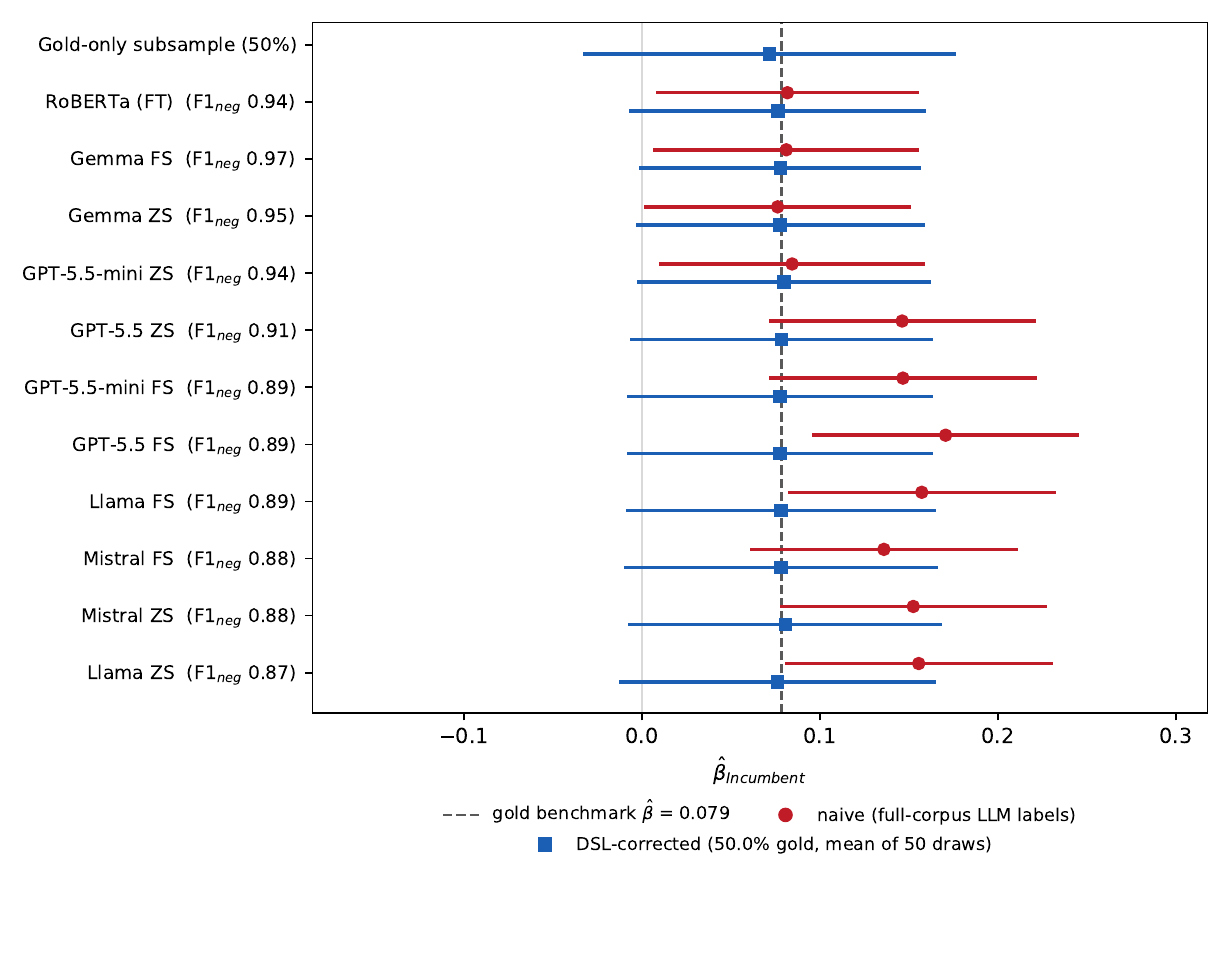}
\end{subfigure}\hfill
\begin{subfigure}[t]{0.435\textwidth}
    \centering
    \caption{100\% gold (DSL = gold regression)}
    \label{fig:foy-dsl-100}
    \includegraphics[width=\linewidth]{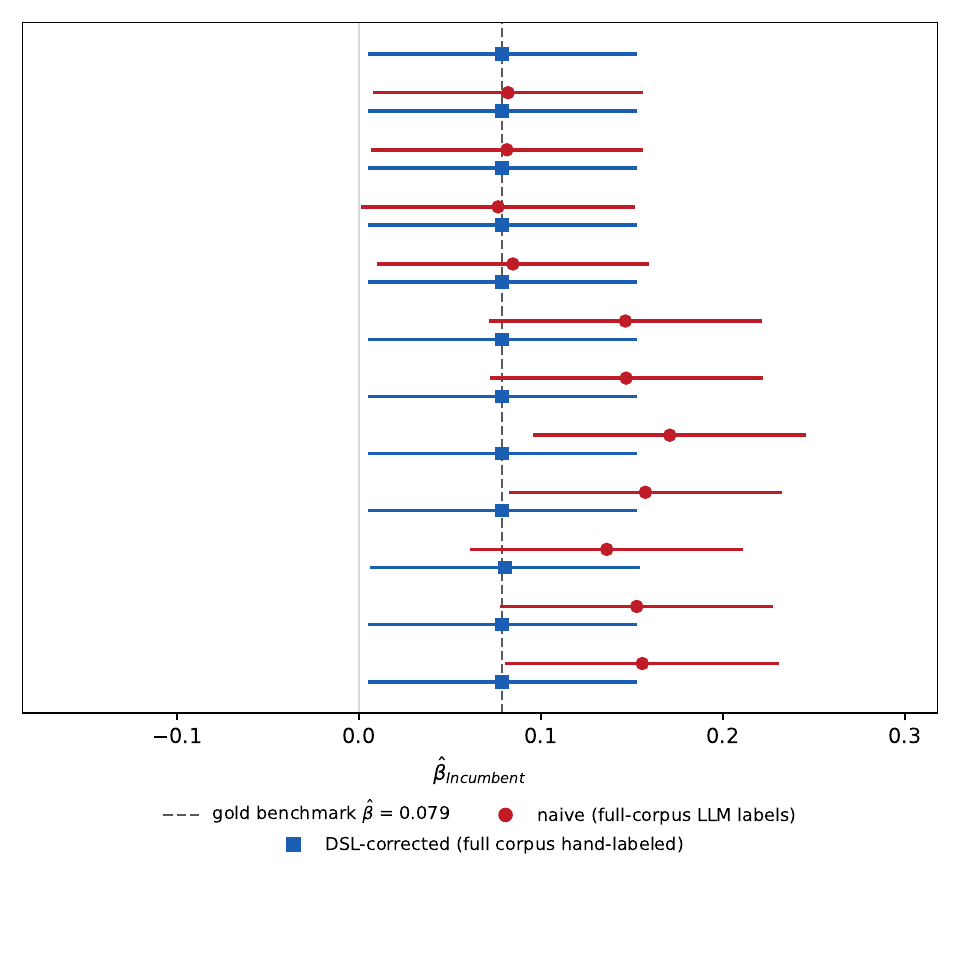}
\end{subfigure}
\vspace{-1.3em}
\caption{Naive (red) versus DSL-corrected (blue) estimates of $\beta_{Incumbent}$ in the \citet{foy2025working} data, by hand-labeling budget. Dashed line and shaded band: gold-label benchmark and its 95\% CI. Configurations ordered by negative-class F1.}
\label{fig:foy-dsl-2x2}
\end{figure}

\subsection{Results: \citet{foy2025working}}

Figure \ref{fig:foy-dsl-2x2} reports the results for the \citeauthor{foy2025working} task. The outcome for negative attack ads is binary, dichotomized from the three-class classifier (negative, contrast and positive). As independent variables, we use an incumbency dummy as the main predictor and the candidate's party and gender as controls. In this task, the best-performing models yield downstream coefficients similar to the gold-label benchmark. RoBERTa matches the best decoder, while other decoders overestimate the effect. DSL corrects these estimates well, but again introduces wider confidence intervals at smaller gold sample sizes. Notice that we predicted these \citeauthor{foy2025working} results at the start of this section: in a task with similar error structures across models, downstream betas are more consistent as model choice barely affects them.

These three results hold some important lessons for downstream inference from LLM-derived measures. First, model choice significantly affects regression coefficients, and the problem is that their effects are unpredictable. In \citeauthor{ballard2023dynamics}, we observe a sign flip and significantly distorted decoder estimates, while the encoder mirrored the benchmark. In \citeauthor{trubowitz2021geopolitical}, the best-performing decoder followed the gold benchmark closely, while RoBERTa and other decoders overstated the effect. In \citeauthor{foy2025working}, both RoBERTa and the best decoders were unbiased with CIs close to the gold benchmark's, and other decoders again overstated the effect. 

Second, in all cases, the DSL correction yields unbiased coefficients, which highlights the importance of the approach when dealing with downstream estimation with LLM-derived measures.\footnote{Recall that DSL requires a hand-labeled subsample drawn as a probability sample, fully at random and with known inclusion probabilities, as in our simulations. A researcher who instead hand-codes a selection of documents (the most recent, the most ambiguous, the easiest to code) for convenience breaks the correction.} DSL did well even in the worst scenarios, such as GPT-5.5-mini zero-shot's performance on the \citeauthor{ballard2023dynamics} data, with 48 percent of rows mislabeled, errors concentrated on the party variable, and a sign-flipped naive $\hat{\beta}$. It also corrected Mistral few-shot's +0.232 overestimation on the same dataset. DSL recovered every one of these estimates and placed them within $\approx$0.02 of the gold-label benchmark. However, it is worth noting that when the DSL sample is small, the estimate can be imprecise, which then affects statistical significance. In fact, all figures show that at 10 or 20 percent of the full sample, DSL-derived CIs are significantly wider than those of both the naive regressions and the gold benchmark. This problem is unlikely to lead to an increase in false positives, so in that sense it is safe, but it may yield a substantially higher number of false negatives (type-II errors), making it more difficult for researchers to establish relationships that do, in fact, exist. For this reason, we introduce a refinement that reduces uncertainty in downstream estimates. 

\subsection{Our Approach: Reducing Variance in Downstream Estimates}

\begin{figure}[!b]
    \centering
    \includegraphics[width=16cm]{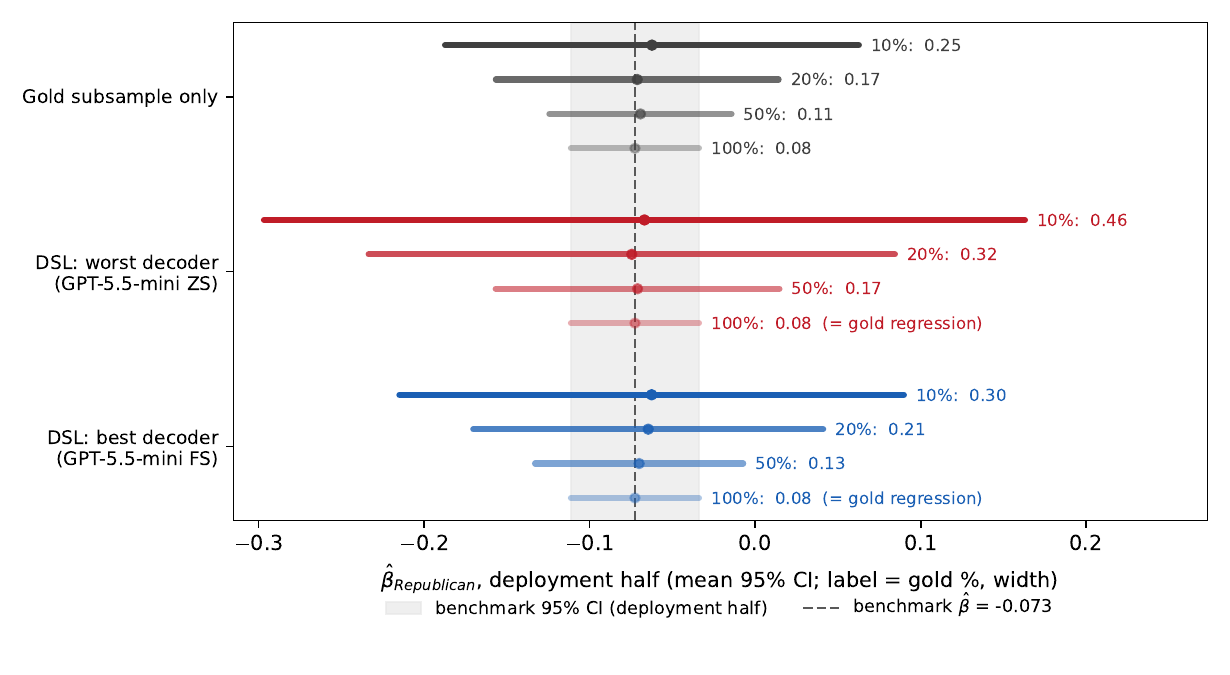}
    \caption{The single-model lottery: DSL-corrected estimates of $\beta_{Republican}$ in the \citet{ballard2023dynamics} data using the worst and best of the ten configurations, against the hand-labeled subsample alone, at three gold budgets. Numbers: mean 95\% CI width. Dashed line and shaded band: gold-label benchmark and its 95\% CI. Which configuration is best or worst is knowable only from a labeled validation sample --- the single-model researcher draws blind.}    
    \label{fig:ballard-lottery}
\end{figure}

The previous subsection shows a key vulnerability in current correction approaches: while DSL works in producing valid downstream coefficients, researchers enter a sort of `precision lottery' around those estimates. The reason is that downstream CIs depend heavily on model choice: a better model for the task will yield narrower CIs than a worse model, even if DSL corrects any estimate and makes it valid. Figure \ref{fig:ballard-lottery} displays this graphically. If a researcher happens to pick the worst decoder for the \citet{ballard2023dynamics} task, GPT-5.5-mini zero-shot, the resulting confidence intervals with DSL at 10 percent of the gold subsample will be 84 percent wider (0.46 vs 0.25), and almost 6 times wider than the \textit{full} gold sample's. At 20 percent of the full sample, DSL's CIs are still 88 percent wider. These numbers improve with the best decoder, GPT-5.5-mini few-shot, but they are still 20 percent wider with the same subsample share (0.30 vs 0.25). Two problems arise. First, researchers cannot know \textit{ex ante} which decoder will produce the best results, as Section 4.6 shows --- hence the term `lottery'. The second issue is the effects of lack of precision. While DSL ensures we obtain valid estimates and reduces the risk of type-I errors (false positives), we are gravely at risk of type-II errors (false negatives), and thus discarding potentially valuable findings.

\begin{figure}[!b]
    \centering
    \includegraphics[width=14cm]{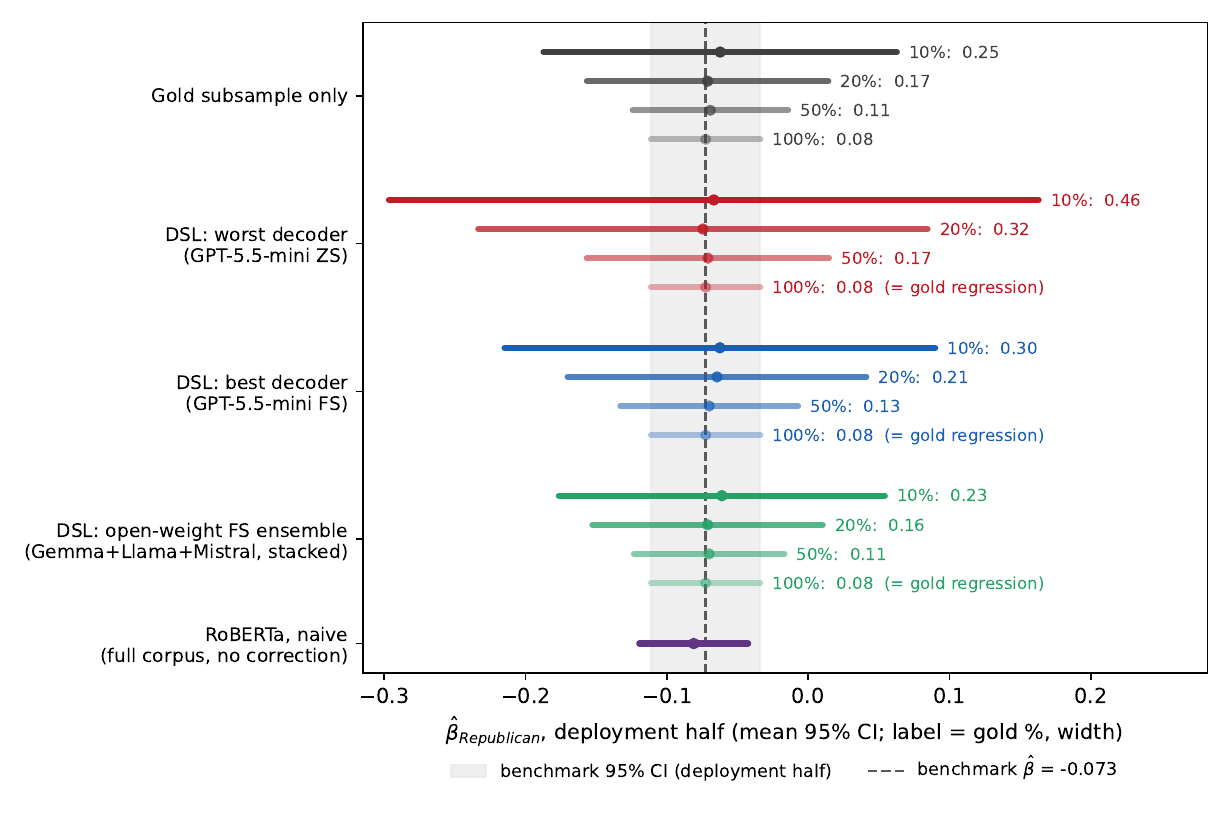}
    \caption{The single-model lottery and the ensemble alternative: DSL-corrected estimates of $\beta_{Republican}$ in the \citet{ballard2023dynamics} data using the worst and best of the ten decoder configurations, the hand-labeled subsample alone, and a stacked ensemble of the three open-weight few-shot configurations \citep[see also][]{timoneda2026bridge} (weights fit on the validation half), at four gold budgets. Bottom row: the naive encoder estimate on the full corpus, uncorrected. Numbers: mean 95\% CI width. Dashed line and shaded band: gold-label benchmark and its 95\% CI.}   
    \label{fig:ballard-lottery-full}
\end{figure}

To address this core inference issue, we suggest two alternative complementary workflows to standard DSL. Both aim to narrow CIs toward the gold benchmark while retaining validity. The first approach is for decoder LLMs and consists in applying DSL over an ensemble of models. To implement and evaluate the ensemble approach, we first split each dataset's full labeled sample into a validation half and a deployment half. On the validation half, we run a logistic regression with the gold label as the dependent variable and the labels from the three open-weight models (Gemma, Llama, and Mistral) as the predictors. Then we apply this fitted model to the deployment half of the data. In it, for each observation, the model takes the three labels it received and returns a single predicted probability that the observation's gold label equals one. These probabilities, one per observation, are the stacked surrogate. This surrogate is the $\hat{y}$ that DSL uses to build its pseudo-outcome. All downstream estimation takes place on the deployment half. There, we simulate a researcher who hand-labels a random 10 or 20 percent of observations (50 and 100 percent serve as comparison points only). Only these drawn gold labels enter the DSL correction. In practice, researchers need not split any data. A single labeled subsample, drawn at random with known inclusion probability, can rank the models, fit the stacking weights, and feed the DSL correction. It is important that the weights be cross-fitted.\footnote{Cross-fitting keeps the surrogate independent of the gold labels used in the correction \citep{egami2024}. With a small stacking model, three coefficients, this costs one line of code: fit the weights on $K - 1$ folds, predict the held-out fold, so every observation's surrogate is out-of-fold.} Figure \ref{fig:ballard-lottery-full} shows the comparison between our ensemble and standard DSL.

To keep costs down, we use only the open-weight models we tested in this book --- Gemma, Llama and Mistral.\footnote{We use the few-shot variants because they tend to perform better on average. That said, researchers are welcome to use a different decoder ensemble, either including some closed-weight alternatives and/or zero-shot prompting variants.} The ensemble yields CIs indistinguishable from the gold subsample and improves on the best decoder's DSL precision by 23 percent at 10 percent of the subsample, 24 percent at 20 percent, and 15 percent at 50 percent. That said, even at 50 percent of gold labels, the ensemble still produces CIs wider than the \textit{full} gold sample's, 0.11 vs 0.08. But the gap is much smaller than standard DSL with either the worst or best decoder. Thus, using a three-model open-weight ensemble addresses two issues. First, it solves the problem of having to select the best model, which is not possible ex ante. Second, although the size of the gain is task-dependent, it can reduce DSL-derived CIs significantly.  

\begin{figure}[!b]
\centering
\begin{subfigure}[t]{0.545\textwidth}
    \centering
    \caption{\citet{foy2025working}}
    \label{fig:foy-lottery}
    \includegraphics[width=\linewidth]{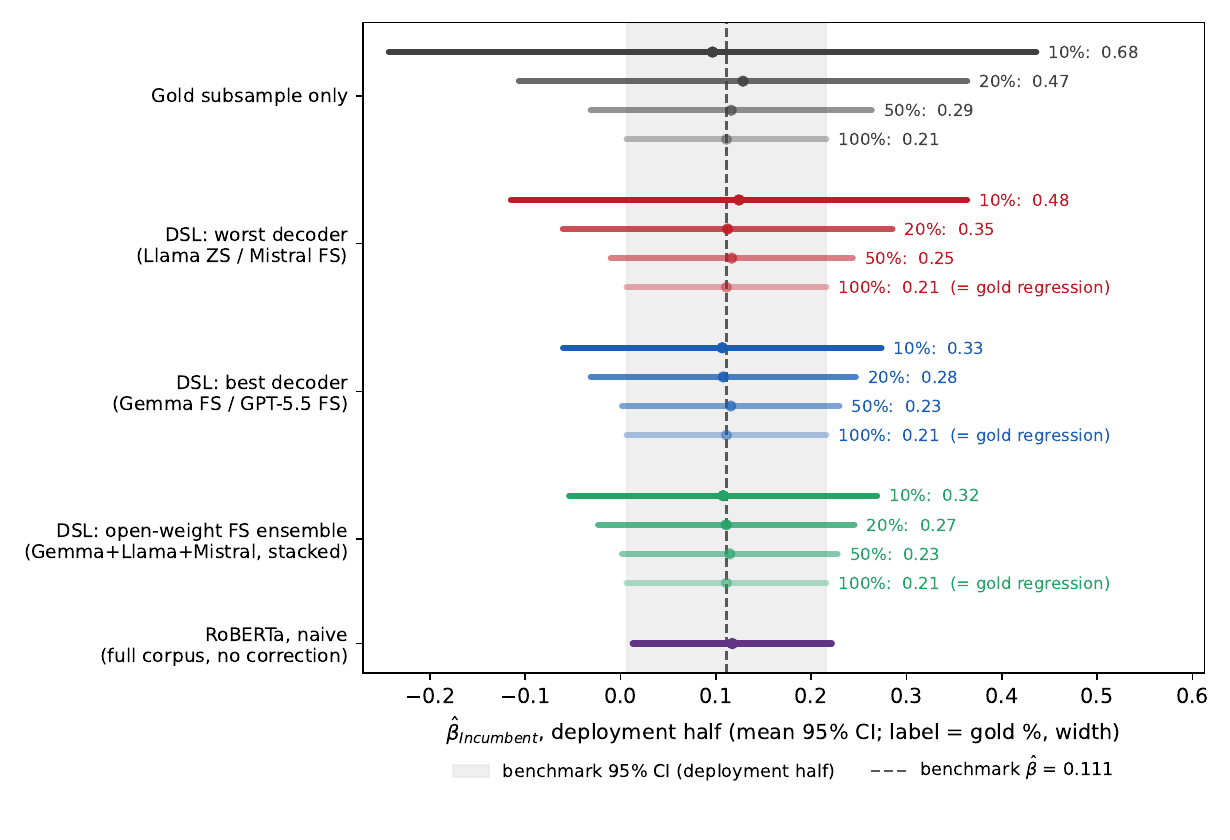}
\end{subfigure}\hfill
\begin{subfigure}[t]{0.435\textwidth}
    \centering
    \caption{\citet{trubowitz2021geopolitical}}
    \label{fig:tru-lottery}
    \includegraphics[width=\linewidth]{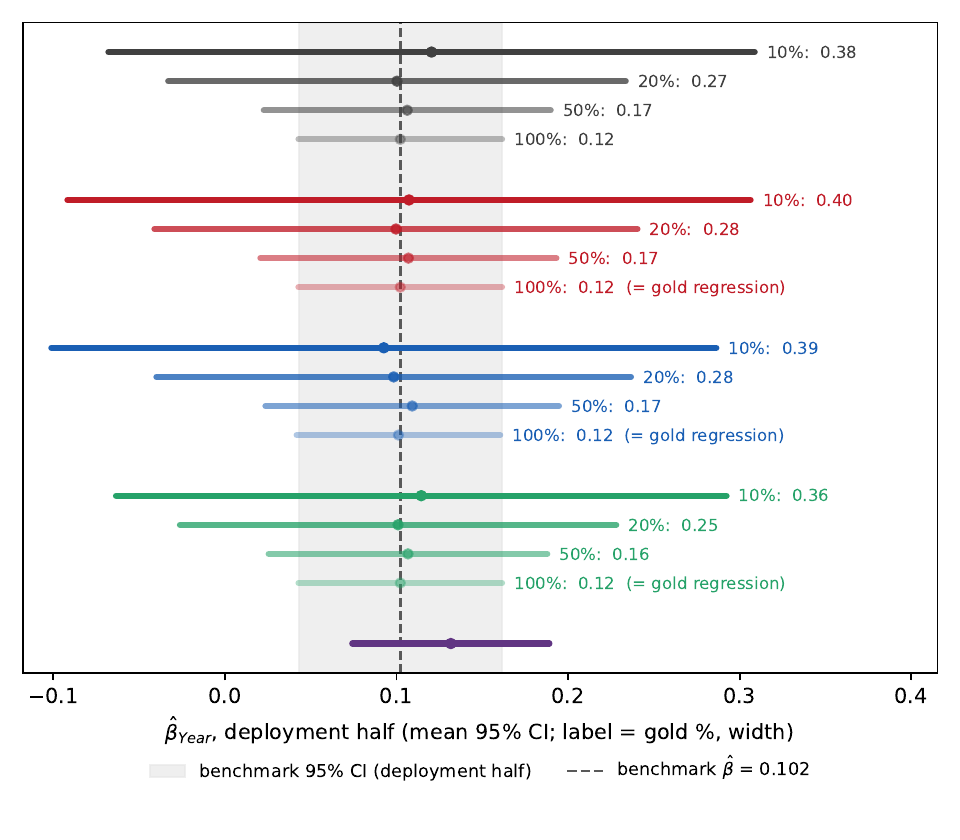}
\end{subfigure}
\caption{The single-model lottery and the ensemble across two further tasks: DSL-corrected estimates using the worst and best of the ten decoder configurations (left label names the worst/best pair as Foy-Sutherland~/~Trubowitz-Watanabe), the hand-labeled subsample alone, and the stacked open-weight few-shot ensemble, at four gold budgets. Bottom rows: the naive encoder estimate on the full corpus, uncorrected. Numbers: mean 95\% CI width. Dashed lines and shaded bands: each task's gold-label benchmark and its 95\% CI.}
\label{fig:lottery-pair}
\end{figure}

Figure \ref{fig:lottery-pair} shows the results for the \citet{foy2025working} and \citet{trubowitz2021geopolitical} tasks. As expected in \citeauthor{foy2025working}, models are the most consistent, which matches the results from this chapter and Section 4.6. Again, the ensemble improves on the best decoder, although the margins are smaller in both tasks. Still, the ensemble is the best \textit{decoder}-based option: it removes the need to select the best model and still yields better performance than the best decoder. 

While the ensemble solution applies only to decoder LLMs, the second uses RoBERTa. We would be remiss if we did not note, from Figures \ref{fig:ballard-lottery-full} and \ref{fig:lottery-pair}, that RoBERTa's results are the most consistently precise of all models. They are also valid, with the estimates in two tasks practically identical to the benchmark and one slightly overstated but within its confidence interval. In fact, based on the tests above, the average decoder needs nearly twice the hand-labeling (18.8 percent of the corpus rather than 10) to match the width of RoBERTa's interval. These results lead us to the second solution: apply a fine-tuned RoBERTa encoder to the full sample, and use the training data benchmark (the same DSL uses) to correct the final downstream estimates. This solution is a supervised-encoder implementation of DSL. 

To explain this second solution, we need to situate the downstream analysis in the broader context of the data. When a researcher has a small amount of data, we highly recommend using a decoder ensemble, as it requires less human annotation, yields valid estimates through the DSL step, and reduces the CIs as much as possible to lower the chance of a type-II error. Fine-tuning a RoBERTa model on the full dataset makes little sense. However, a researcher may have \textit{a lot of texts} and the decision is between (1) using a decoder and applying DSL with a small validation set or (2) building a full training set for fine-tuning RoBERTa (or a decoder with QLoRA). Here is where we think fine-tuning RoBERTa still has an edge that is difficult to surpass. Our tests have shown that RoBERTa generally produces strong estimates, and the CIs mirror those of the full gold-label benchmark because that is the sample it uses for fine-tuning. The second solution, therefore, is to run DSL with RoBERTa as the surrogate: the researcher applies the fine-tuned classifier to the full corpus, producing a predicted label for every document. We use the out-of-fold predictions for the training observations themselves, so the surrogate never sees its own gold labels. The training set then serves as the gold subsample in the DSL step. The corrected estimates are unbiased and the CIs are also as narrow as in the gold-benchmark regression.\footnote{The width of the corrected interval is anchored to the gold-benchmark sample, not to the size of the corpus, but it is scaled by the quality of the encoder: it is narrower than the gold-only interval whenever the encoder's per-document error falls below the outcome's residual variance, and wider when it does not. This break-even condition can be computed  from the validation sample before a single document of the full corpus is labeled. Note that this approach, incidentally, also helps prevent type-I errors emerging from large sample sizes common to text sources.}

In all, one way to describe the challenges of using LLMs as annotators is that they put researchers between a rock and a hard place: either they use an LLM blindly and risk biased downstream estimates, or they use DSL with a small gold-labeled sample to correct for bias but obtain imprecise estimates. The latter is certainly preferable to the former, but neither is ideal. The main takeaway, for us, is that we \textit{cannot fully turn our backs on building a training set} if we want to ensure that both betas and CIs are valid for downstream inference. The remaining decision point for a researcher is \textit{how much data} to label. If there are clear resource constraints, labeling a small set and applying DSL with an ensemble of open-weight LLMs is a good option. The costs are somewhat wider intervals than a full gold sample would give, and a weaker basis for judging LLM performance on a specific task. If resources are not a problem, a larger labeled set unlocks the fine-tuning options through encoders and QLoRA, which yield the narrowest valid CIs our tests have produced. 
\section{Transparency and Disclosure}

More than a methodological issue, transparency is a normative one. As a discipline, we have adopted general standards of disclosure and data availability, such as APSA's Data Access and Research Transparency (DA-RT) initiative. And while there have been improvements in this area, there is still much work to do \citep{grossman2021can,rainey2025data}. This is not limited to research using LLMs, although LLMs pose new challenges \citep[see][]{grossmann2023ai,abdurahman2025primer,feuerriegel2026reporting}. We echo these calls. The disclosure standards already in place at top journals are the floor, and yet they have not closed the reporting gap, least of all for LLM-based work.

\begin{table}[!h]
\centering
\caption{Information to Disclose When Using LLMs}
\label{tab:llm_disclosure}
\begin{tabular}{p{0.25\linewidth} p{0.65\linewidth}}
\toprule
\textbf{Disclosure item} & \textbf{Information to report} \\
\midrule
Model identification & 
Name, version, and repository identifier of the LLM or language model used, including whether it is a base, fine-tuned, or otherwise modified model. \\

Model modifications & 
Any modifications applied to the model, including additional pre-training, fine-tuning, instruction tuning, adapters, prompting frameworks, or other architectural or procedural changes. \\

Prompt(s) & Full text of the final prompt, including system message, category definitions, few-shot examples, and required output format; the number of repeated runs; and a summary of prompt variants tested. \\

Training or fine-tuning data & 
Description of any additional data used to further pre-train, fine-tune, or adapt the model, including data source, size, selection criteria, preprocessing, and relevant exclusions. \\

Hyperparameters & 
Relevant training, fine-tuning, or inference hyperparameters, such as learning rate, batch size, number of epochs, optimizer, temperature, top-$p$, maximum tokens, and random seeds where applicable. \\

Computational equipment & 
Hardware used for training, fine-tuning, or inference, including GPU or CPU type, number of devices, memory, and other relevant computational specifications. \\

External computing services & 
Any external computing service used, such as cloud platforms, hosted APIs, or rented servers. For example, researchers should report if a RoBERTa model was fine-tuned using a datacrunch.io server. \\

Access and availability & 
Whether the model, code, prompts, fine-tuned weights, training data, or inference outputs are publicly available, restricted, or unavailable, and how they can be accessed where possible. \\

Replicability notes & 
Any additional details needed to replicate the use of the model, including software libraries, package versions, API settings, dates of access, and limitations on exact replication. \\
\bottomrule
\end{tabular}
\end{table}

With LLMs, the key issues are what to disclose, what to make available, and how. In terms of disclosure, researchers must first disclose model name, exact version, hyperparameters, and modifications, meaning any further pre-training, fine-tuning, quantization, or adapters applied to the released checkpoint. Computational equipment details and any external computing service or API used must also be disclosed. We provide a complete overview of disclosure criteria in Table \ref{tab:llm_disclosure}. For example, a researcher who fine-tunes or further trains a RoBERTa model on Verda or AWS should report the provider and instance. Any additional training data should also be described in detail (see Table \ref{tab:llm_reporting_matrix}). The date of access is also essential, since models change or get deprecated. 

\begin{sidewaystable}[htbp]
\centering
\scriptsize
\caption{Examples of different LLM applications}
\label{tab:llm_reporting_matrix}
\begin{tabularx}{\linewidth}{p{0.16\linewidth}YYYY}
\toprule
\textbf{Reporting criterion} 
& \textbf{RoBERTa} 
& \textbf{Further trained RoBERTa} 
& \textbf{Mistral} 
& \textbf{Mistral + QLoRA} \\
\midrule

Model identification 
& RoBERTa-base, Hugging Face implementation, version accessed on date of analysis. 
& RoBERTa-base further pre-trained by the researchers before task-specific fine-tuning. 
& Mistral-7B, decoder-only instruction or base model, version accessed on date of analysis. 
& Mistral-7B adapted with QLoRA weights for the target task. \\

Model modifications 
& No architectural modifications; model fine-tuned for the classification task. 
& Additional domain-adaptive pre-training followed by task-specific fine-tuning. 
& No weight updates; model used through prompting for inference. 
& QLoRA adapters added to selected attention layers; base model weights frozen. \\

Training or fine-tuning data 
& Task-specific labeled dataset, e.g., 8,000 training observations and 2,000 validation observations. 
& Additional unlabeled domain corpus, e.g., 50 million tokens from Judicial Review corpus, plus the same labeled task dataset. 
& No additional training data; prompts applied directly to the evaluation dataset. 
& Task-specific instruction-style dataset, e.g., 5,000 prompt-response examples from C-SPAN interviews. \\

Hyperparameters 
& Learning rate $2 \times 10^{-5}$; batch size 16; 3 epochs; AdamW optimizer; seed 42. 
& Pre-training learning rate $1 \times 10^{-5}$; fine-tuning learning rate $2 \times 10^{-5}$; batch size 16; 3 epochs. 
& Temperature 0.2; top-$p$ 0.9; maximum output length 512 tokens; fixed prompt template. 
& QLoRA rank 16; QLoRA alpha 32; dropout 0.05; learning rate $2 \times 10^{-4}$; 3 epochs; temperature 0.2 at inference. \\

Computational equipment 
& Single NVIDIA RTX A6000 GPU with 48GB memory. 
& Single NVIDIA A100 GPU with 80GB memory each for further pre-training; one RTX A6000 GPU with 48GB for fine-tuning. 
& Inference run on hosted GPU service or local NVIDIA A100 GPU. 
& Fine-tuning run on one NVIDIA A100 GPU with 40GB memory; inference on the same hardware. \\

External computing service 
& No external service used, or alternatively: fine-tuned on a rented datacrunch.io (verda.com) server. 
& Further pre-training and fine-tuning conducted on datacrunch.io (verda.com) GPU instances. 
& Model accessed through a hosted inference endpoint or local deployment. 
& Base model and QLoRA fine-tuning conducted on rented cloud GPU infrastructure. \\

Access and availability 
& Code and fine-tuned weights made available in a public repository, subject to data restrictions. 
& Further pre-trained weights shared where licensing permits; additional corpus described but not fully released if restricted. 
& Prompt templates, model name, and inference settings released; model weights available from provider or repository. 
& QLoRA adapter weights, prompts, and fine-tuning scripts released; base model obtained separately according to its license. \\

Replicability notes 
& Report software versions, including Python, PyTorch, Transformers, CUDA, and random seed. 
& Report additional pre-training procedure, corpus construction, checkpoints, software versions, and random seeds. 
& Report prompt template, system message, inference date, decoding parameters, and API or model version. 
& Report QLoRA configuration, adapter checkpoints, base model version, prompt template, inference settings, and software versions. \\

\bottomrule
\end{tabularx}
\end{sidewaystable}

For availability, the default should be to release everything the license and ethics allow. We are in an age of low data storage costs and ample repository availability. Setting aside the common exceptions (confidentiality, security, informed-consent limits), researchers should make the following readily available: code, prompts, codebooks, training and validation data, analysis data, and trained models.\footnote{Availability requires a certain degree of trust in current repositories and platforms. GitHub and Hugging Face, for example, are invaluable resources for researchers, but repositories hosted there might change after publication or without notice; the platforms themselves might change or disappear. Ideally, all archived versions should be assigned a DOI or a similar persistent identifier. All published material should link back to that identifier, preserving the dated version of the reported material, while allowing the researcher to continue maintaining a working repository.} In what follows we describe how each should be made available.

\subsection{Code Scripts}

Regardless of model, a general rule is that all code should be self-contained, annotated, and easily interpretable. By self-contained, we mean that researchers should be able to run the complete workflow with minimal changes. If fine-tuning an encoder model, the code should point to the correct repositories to access the same exact model version,\footnote{See the Trained Models section for model availability.} and the correct training data using relative paths. If using a fine-tuned model, the code should point to the correct repositories or folders that store the model, the tokenizer and the data. When using a decoder model, the code should include either the API call used to access it or the correct repository that stores it. The exception is secret API keys, which should never be shared nor hardcoded into replication scripts. The code does need to indicate where to place the API key. 

For better readability, we encourage researchers to thoroughly and clearly annotate the code they share. This means describing the overall goal of the script, the libraries and their versions, the functions and each of their components, and the steps the script executes. When possible, objects should follow a consistent and intuitive naming convention; the script should signpost final outputs and include relevant information about the computational resources used and run time. Finally, a self-contained workflow should also include a README file with the order of execution and other relevant information that will make our material easier to access and replicate. 

\subsection{Prompts}

In earlier chapters we have discussed and tested how varying prompt structure can influence output. In Section 5.3, we presented a recommended template for researchers to apply in their prompting tasks. We believe that using a template can help standardize reporting practices. While it is unrealistic to ask researchers to report all attempted prompts, researchers should report how they arrived at the final prompt, what aspects of the prompt were changed throughout testing,\footnote{Examples are the roles defined for the LLM, the order of categories, the examples used, etc. If the researcher systematically evaluated different prompt variants, then these should be reported as well.} and how those changes affected the final output. The prompt should detail the structure of the responses requested, such as `respond with either 0 or 1', `respond in one sentence', `provide a summary no longer than 300 words', or `provide the reasoning behind your answer'. While none of these reporting recommendations prevent prompt-hacking, they provide useful information for future researchers looking to replicate work and planning their own research.

\subsection{Codebooks}

Researchers need at least an expert-coded training or validation set to deploy any LLM. Annotation and validation quality relies on the quality of our training and validation sets. The codebooks used to produce these labels must be released, and researchers should document the instructions provided to coders. Any changes to the codebook should also be reported, as well as any lessons learned from the labeling task.\footnote{Labeling data is an iterative process that often requires multiple rounds of training and adjusting definitions. While the discipline values deductive approaches to labeling data (i.e., first we define the concept and then we identify it in the data), it is not uncommon to find variation of the concept or different categories altogether as we carry out this exercise inductively. We encourage researchers to report this as well.} This reporting exercise serves many purposes. For one, it provides a benchmark for what we expect to capture in our task, allowing us to determine whether human and LLM labels capture concepts accurately. It also enriches the discipline, as codebooks and the processes to develop them are rarely reported, which deprives other researchers of valuable insights and lessons. Finally, codebook instructions are rarely copied verbatim into prompts. Releasing both lets readers see the choices made in translating instructions written for human coders into text built for the LLM. 

\subsection{Training sets and validation sets}

In the social sciences, we pride ourselves on meticulous and theory-driven approaches to operationalize concepts. In our experience, the discipline is still a long way from meeting the standards it has largely agreed to. Training and validation sets must clearly identify two elements, the text to classify and its label.\footnote{The process to obtain test, training, and validation sets is the same. Our recommendations apply to all three.} Additionally, training sets must be accompanied by documentation naming the column that carries the concept and defining every label value. Thus, if we were to classify a text as either positive (1), neutral (0), or negative (-1), it should be clear to other users what each number in the classification column means. If the training set requires merges to be usable, for example when the text lives in a separate file, then matching IDs and the merge code must be provided. Beyond transparency, clearly and cleanly reported training sets are a public good that could eventually allow for the creation of disciplinary benchmarks for what \textit{good} performance means on the tasks we care about. 

\subsection{Training data}

Chapters 3 and 4 showed that we can further adapt encoder and decoder LLMs to specific domains. Encoders can be further trained using raw, unstructured text data. If permissions allow, we encourage researchers to make training data available to the public, including metadata like its source, the time frame it covers, and a snippet of the data. When providing access to the data, lightweight formats such as .txt are preferable, and it is always good practice to provide the code if the data requires merging or parsing. For decoders, on the other hand, we can adapt them with instruction-style prompt--response pairs, or with the ranked preference data used in RLHF. Further pre-training carries prohibitive costs, so efforts focus on these two post-training steps. In this book, QLoRA fine-tuning is exactly such supervised adaptation, and its training pairs deserve the same disclosure. For ranked question-answer pairs, in addition to the source and date, researchers should provide an explanation of how the questions and answers were matched, how the rankings were done, and the code used to structure the raw data. 

\subsection{Trained models}

In the making of this book, we ran more than 1,500 models and fine-tuned over 1,200. Not all the models can be made available at Hugging Face's model repository, but researchers should publish the final versions of their models on the site. Transparency, we believe, means providing not only the complete workflow, but also the results of the intermediate steps. Researchers often share the code to fine-tune a model, but they should also share the final model used in their analysis. First, this allows other researchers to directly compare output when either fine-tuning the model again or deploying it, which is fundamental for replicability. Second, it becomes a public good whose generalizability other researchers can test by deploying it on new corpora. For models to be useful to other users, researchers need to include complete model cards, all the information we outline in this section, and clearly marked labels that match the output of the models. Note that this can, and should, be done for encoder and decoder models.   

We have provided a guide to transparency and tried to follow it in the development of this book. The repository accompanying this book contains the code, prompts, training data, and models behind every result (see {\small \bookrepo}). Transparency is what makes the rest of this chapter possible. Readers can only test replicability, reproducibility, and robustness against what authors disclose.

\section{Conclusion}

This chapter took the reproducibility and transparency debate on its own terms and attempted to give it a constructive turn. We believe the stochastic nature of LLMs does not mean that LLM research is not replicable, reproducible or robust. We proposed a general switch to probabilistic RT, which requires moving away from determinism into the analysis of distributions of results. We showed that the familiar tools of the discipline, from repeated cross-validation to disclosure norms, extend naturally to it. For the robustness of results across different LLMs, we argued that what matters is \textit{where} models differ, and that the error-structure analysis tells researchers when the best model suffices and when an LLM ensemble approach is necessary. For downstream inference, DSL guarantees validity but can lower precision, and we proposed an ensemble approach that yields narrower and valid CIs. Transparency, finally, is what makes everything testable, and we have tried to hold this book to the standard it proposes. The deeper lesson runs through every section: the age of decoder LLMs has not abolished labeled data, but rather transformed the \textit{need} for human-coded labels. It has changed the gold sample's job, from fine-tuning the model to validating inference. A strong training and validation set is still one of the best investments an applied researcher can make.

\clearpage
\bibliography{book_literature}
\bibliographystyle{apsr}

\end{document}